\documentclass[journal]{IEEEtran}
\usepackage[utf8]{inputenc}
\usepackage{mathtools,lipsum,cuted}
\usepackage{amsmath,amsfonts}
\usepackage{enumitem}
\usepackage{algorithmic}
\usepackage{array}
\usepackage{stfloats}
\usepackage{url}
\usepackage{bm}
\usepackage{verbatim}
\usepackage{graphicx}
\usepackage[table]{xcolor}
\usepackage{enumerate}
\usepackage[table]{xcolor}
\usepackage{rotating}
\usepackage{multirow}
\usepackage{makecell}
\usepackage{booktabs}
\usepackage{multirow}
\usepackage{amssymb}   
\usepackage[caption=false,font=footnotesize]{subfig}
\usepackage{multirow}
\usepackage{makecell}
\usepackage{booktabs}
\usepackage[table]{xcolor}
\usepackage{hyperref}

\newcommand{\best}[1]{\cellcolor[RGB]{248,226,214}#1}

\newcommand{\std}[1]{\textbf{#1}}
\definecolor{pav1}{RGB}{0, 0, 128}
\definecolor{pav2}{RGB}{0, 31, 251}
\definecolor{pav3}{RGB}{47, 142, 252}
\definecolor{pav4}{RGB}{130, 254, 230}
\definecolor{pav5}{RGB}{172, 253, 144}
\definecolor{pav6}{RGB}{239, 253, 81}
\definecolor{pav7}{RGB}{245, 160, 47}
\definecolor{pav8}{RGB}{239, 58, 28}
\definecolor{pav9}{RGB}{129, 6, 1}

\definecolor{h13c01}{RGB}{255, 255, 5}
\definecolor{h13c02}{RGB}{144, 238, 145}
\definecolor{h13c03}{RGB}{75, 0, 130}
\definecolor{h13c04}{RGB}{255, 165, 3}
\definecolor{h13c05}{RGB}{255, 192, 203}
\definecolor{h13c06}{RGB}{205, 133, 63}
\definecolor{h13c07}{RGB}{220, 20, 60}
\definecolor{h13c08}{RGB}{3, 255, 255}
\definecolor{h13c09}{RGB}{32, 144, 255}
\definecolor{h13c10}{RGB}{112, 129, 145}
\definecolor{h13c11}{RGB}{0, 0, 255}
\definecolor{h13c12}{RGB}{255, 0, 0}
\definecolor{h13c13}{RGB}{3, 139, 139}
\definecolor{h13c14}{RGB}{128, 128, 0}
\definecolor{h13c15}{RGB}{0, 128, 1}
\definecolor{h18c01}{RGB}{51, 205, 52}
\definecolor{h18c02}{RGB}{173, 255, 48}
\definecolor{h18c03}{RGB}{0, 128, 129}
\definecolor{h18c04}{RGB}{35, 139, 35}
\definecolor{h18c05}{RGB}{47, 79, 78}
\definecolor{h18c06}{RGB}{139, 69, 18}
\definecolor{h18c07}{RGB}{3, 255, 255}
\definecolor{h18c08}{RGB}{255, 255, 255}
\definecolor{h18c09}{RGB}{211, 211, 211}
\definecolor{h18c10}{RGB}{254, 1, 0}
\definecolor{h18c11}{RGB}{169, 169, 169}
\definecolor{h18c12}{RGB}{105, 105, 105}
\definecolor{h18c13}{RGB}{139, 0, 0}
\definecolor{h18c14}{RGB}{200, 101, 1}
\definecolor{h18c15}{RGB}{255, 165, 0}
\definecolor{h18c16}{RGB}{255, 255, 5}
\definecolor{h18c17}{RGB}{218, 165, 33}
\definecolor{h18c18}{RGB}{255, 0, 254}
\definecolor{h18c19}{RGB}{0, 0, 254}
\definecolor{h18c20}{RGB}{64, 224, 208}

\definecolor{bestshade}{RGB}{248,226,214}

\renewcommand\eqref[1]{(\autoref{#1})}
\usepackage{mathtools,lipsum,cuted}
\usepackage{amsmath}     
\usepackage{amssymb}                       
\usepackage{mathrsfs}    
\usepackage{upgreek}
\usepackage{bm}     
\usepackage{authblk}
\usepackage{setspace}
\usepackage{graphicx}
\usepackage{amsfonts,amssymb}
\usepackage{caption}
\graphicspath{ {./figures/} }
\graphicspath{ {./Author/} }
\usepackage{amsmath}
\usepackage{cuted}
\usepackage{lineno}
\usepackage{colortbl}
\usepackage{bbding}
\usepackage{multirow}
\usepackage{makecell}
\usepackage{blindtext}
\usepackage{xcolor}
\usepackage{color}
\usepackage{setspace}
\usepackage{makecell}
\usepackage{threeparttable}
\usepackage{hyperref}
\usepackage[section]{placeins}
\usepackage{hhline}
\usepackage{threeparttable}
\usepackage{tikz}
\usepackage{diagbox}
\usepackage{algorithm,algorithmic}

\hypersetup{
	colorlinks=true,
	linkcolor=cyan,
	filecolor=blue,      
	urlcolor=black,
	citecolor=green,
}
\DeclareMathAlphabet\mathbfcal{OMS}{cmsy}{b}{n}
\begin{document}

\title{Token Clustering and Semantic Sequence Mamba for Hyperspectral Image Classification}

\author{
Yimin Zhu, Mahmood Elahi, Lincoln Linlin Xu, ~\IEEEmembership{Member,~IEEE} 
\thanks{Yimin Zhu is with the Department of Geomatics Engineering, University of Calgary, Canada (email: yimin.zhu@ucalgary.ca)}
\thanks{Mahmood Elahi is with the Department of Electrical and Software Engineering, University of Calgary, Canada (email: mahmood.elahi@ucalgary.ca)}
\thanks{Corresponding author Lincoln Linlin Xu is also with the Department of Geomatics Engineering, University of Calgary, Canada (email: lincoln.xu@ucalgary.ca)}
\thanks{This work was supported by the Natural
Sciences and Engineering Research Council of Canada (NSERC) under Grant RGPIN-2019-06744.}
}

\markboth{Journal of \LaTeX\ Class Files}%
{Shell \MakeLowercase{\textit{et al.}}: Bare Demo of IEEEtran.cls for Journals}

\maketitle

\begin{abstract}
Although hyperspectral images (HSIs) provide rich spectral-spatial information, accurate pixel-level classification remains challenging because of spectral-spatial heterogeneity and complex spatial structures. Existing vision state-space models (Mamba) typically construct sequences according to predefined spatial neighborhoods, without explicitly accounting for semantic similarity or spatial non-stationarity. To address this limitation, we propose Token Clustering and Semantic Sequence Mamba (STMamba), which organizes sparse tokens into semantically coherent sequences for hyperspectral image classification with the following features. First, at the macro level, a hierarchical encoder–decoder progressively selects semantic tokens with the Token Clustering Module (TCM) and restores dense features using a parameter-free Cross-scale Neighborhood Attention (CNA) Upsampler. Second, at the micro level, TCM first identifies representative cluster centers through density-aware clustering and estimates soft memberships based on feature similarity. A quadtree-based dynamic selection strategy then retains sparse and spatially distributed tokens from each semantic cluster, forming coherent semantic-token sequences while reducing redundant pixel-wise representations. Third, parallel Spatial and Spectral Semantic-wise Sequencing Mamba (SWSM) modules capture complementary long-range spatial and spectral dependencies within homogeneous semantic-token sequences while suppressing irrelevant interactions across heterogeneous regions.
Experimental results on three large-scale benchmark datasets demonstrate that STMamba outperforms the state-of-the-art methods with respect to quantitative and qualitative results.

\end{abstract}


\IEEEpeerreviewmaketitle
\section{Introduction}

\IEEEPARstart{H}yperspectral images provide rich spatial--spectral information for distinguishing spectrally similar materials and support applications such as environmental monitoring \cite{Alakian2024} geologic mapping \cite{HAJAJ2024101218}, and agricultural disaster response \cite{AKHYAR2024112067}. However, hyperspectral image classification (HSIC) remains challenging because spectral--spatial heterogeneity, mixed pixels, low signal-to-noise ratio, and illumination variations produce substantial spectral variability and complex spatial structures \cite{9583297,han2025subpixel}. Effective HSIC therefore requires representative feature extraction, spatial--spectral dependency modeling, and efficient long-range contextual interaction.

\begin{figure*}[!t]
\centering
\includegraphics[width=\textwidth]{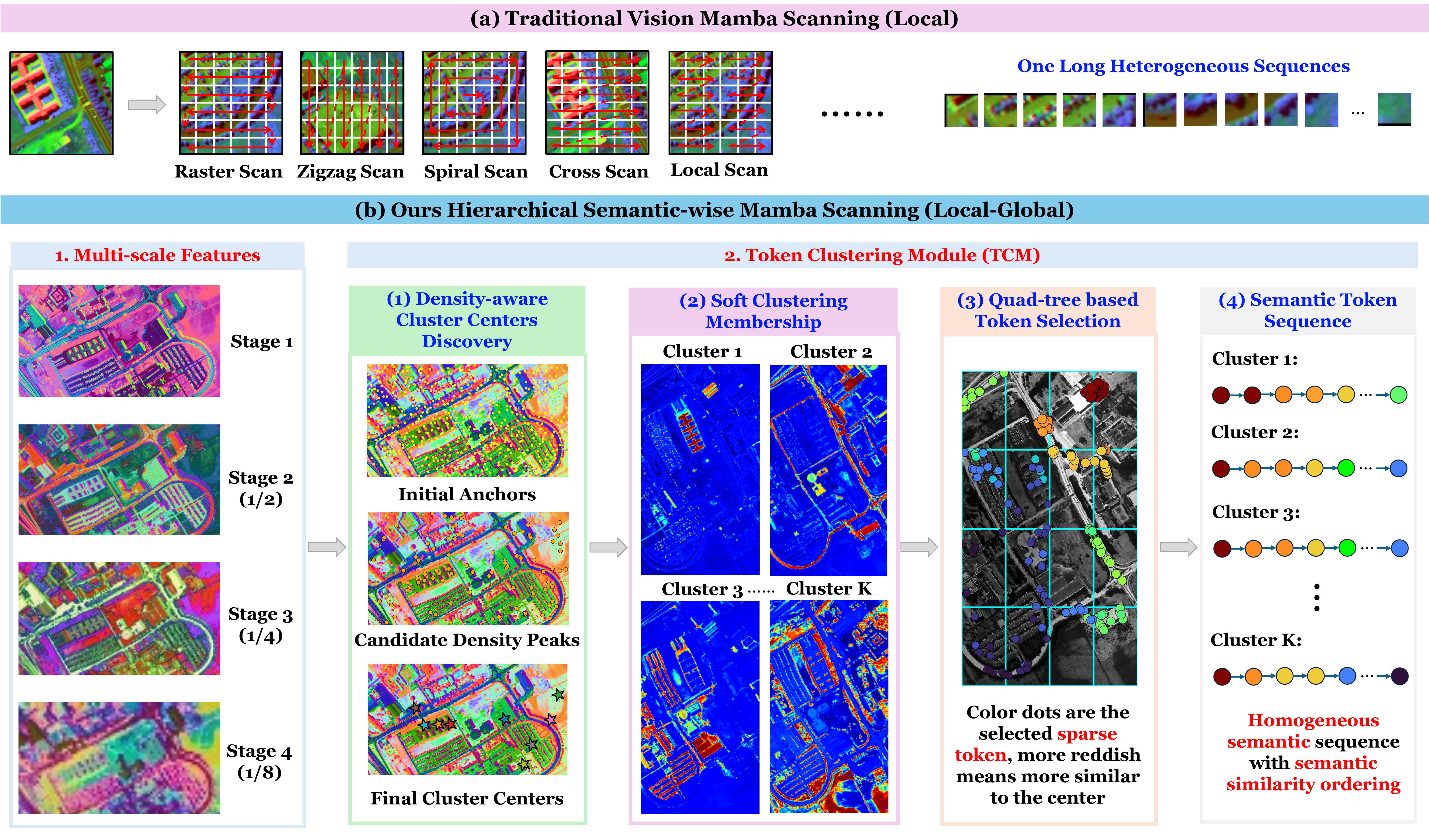}
\caption{\textbf{(a) Traditional Vision Mamba Token Sequencing} organizes an entire patch or image using predefined scanning paths despite heterogeneous spatial patterns \cite{wang2025mambahsiplus}. \textbf{(b) Our method} adaptively selects sparse representative tokens through density-aware clustering and organizes them into homogeneous semantic sequences for Spatial and Spectral Semantic-wise Sequencing Mamba (SWSM).}
\label{Idea_fig}
\end{figure*}

Traditional machine-learning methods primarily model pixel-wise spectral relationships and have limited ability to represent complex spatial--spectral structures. CNN- and Transformer-based approaches improve spatial--spectral modeling, but patch-based processing restricts the field of view and introduces redundant computation \cite{11563901}, while self-attention has quadratic complexity with respect to token length. With the emergence of large-scale HSIs \cite{ji2025patchout}, patch-free encoder--decoder frameworks have been developed to efficiently exploit global spatial information \cite{zheng2020fpga,ji2025patchout}. Similar strategies have recently been adopted by Mamba-based HSIC models \cite{11563901,10604894}. Nevertheless, most approaches still construct visual tokens using regular spatial grids and treat them approximately equally, without explicitly considering geospatial non-stationarity and abrupt semantic transitions between neighboring pixels \cite{li2026pixel}. As illustrated in \autoref{Idea_fig}(a), predefined scanning may consequently place heterogeneous regions within the same sequence and introduce unnecessary interactions between semantically unrelated pixels. Moreover, sparse benchmark annotations primarily evaluate labeled interior regions and provide limited assessment of semantic boundary preservation. These issues motivate the construction of sparse, representative, and semantically meaningful tokens for whole-image HSIC.

Spatial--spectral decoupling provides another way to address HSI heterogeneity. Existing dual-branch, decomposition, and aggregation networks separately model different feature domains \cite{11396057,11421023}. Sub-pixel decomposition can further expose homogeneous components within mixed observations \cite{10570241}, while spatial--spectral decoupling has also been implemented through auxiliary Transformer branches \cite{9684381}, domain-specific convolution kernels \cite{8061020,10144309}, and feature-token representations \cite{11313493,FAN2025101547}. However, these methods generally do not explicitly discover homogeneous semantic regions before dependency modeling. Similarity-based token grouping, for example, may still depend on a predefined center pixel and consequently have limited adaptability. As shown in \autoref{Idea_fig}(b), we instead associate pixels with representative semantic centers before separately modeling spatial and spectral dependencies.

Mamba provides an attractive mechanism for long-range modeling because its computational complexity scales linearly with sequence length, i.e., $\mathcal{O}(L)$, compared with $\mathcal{O}(L^2)$ for standard self-attention \cite{ma2024u}. Existing HSI Mamba models, including MambaHSI \cite{10604894}, GraphMamba \cite{10746459}, DBMGNet \cite{10976709}, and MiM \cite{zhou2025mamba}, mainly operate on densely constructed sequential or graph tokens. Recent studies have explored adaptive or sparse token selection \cite{wang2024graph}. Xu et al. \cite{11075710}, for example, select and sort deformable tokens according to their similarity to a central pixel and anchor band, while token-prioritization scores are adopted in \cite{ahmad2025hybrid}. Although these strategies reduce redundant tokens, token selection and permutation are not jointly determined by semantic structure. Therefore, the key problem is not only to make Mamba scanning sparse or dynamic, but to construct semantically coherent sequences in which representative tokens from homogeneous regions can interact over long distances.

To address these issues, we propose Token Clustering and Semantic Sequence Mamba (STMamba), and reformulate Mamba sequence construction from geometry-defined scanning to semantic-membership-defined sparse sequencing, with following features:

1. At the macro level, a hierarchical encoder--decoder progressively extracts semantic tokens using a Token Clustering Module (TCM) and restores dense representations through a parameter-free Cross-scale Neighborhood Attention (CNA) Upsampler.

2. At the micro level, TCM identifies representative centers using density-aware clustering and estimates soft semantic memberships from feature similarity. A quadtree-based dynamic strategy then selects sparse and spatially distributed tokens from each semantic cluster, producing coherent semantic-token sequences while reducing redundant pixel-wise representations.

3. Parallel Spatial and Spectral Semantic-wise Sequencing Mamba (SWSM) modules model long-range spatial and spectral dependencies within homogeneous semantic-token sequences, while reducing irrelevant interactions across heterogeneous regions.

The remainder of this article is organized as follows. \autoref{related_work} reviews related work, \autoref{method_part} presents the proposed method, \autoref{exp} reports the experimental results, \autoref{discussion} provides further discussion, and \autoref{conclusion} concludes the article.

\section{Related Work} \label{related_work}

\subsection{Patch-free Network for HSIC}

Patch-free HSIC methods process the entire hyperspectral image to capture global context while avoiding repeated computation over overlapping patches~\cite{woo2023convnext,zhao2025classification}. For example, PBiNet adopts separate spatial and semantic branches to preserve fine details and high-level semantics~\cite{liu2021patch}; Spe-TL models spectral variations by treating HSI bands as a sequence of images~\cite{sun2022video}; and PatchOut combines Transformer-based long-range modeling with convolutional local feature extraction in an encoder--decoder architecture~\cite{ji2025patchout}. For large-scale imagery, LS3EU-Net++ further employs a lightweight U-Net++ with spatial--spectral attention to generate full-resolution predictions~\cite{zhao2025classification}. Despite their efficiency, these methods generally extract features from regular spatial grids without explicitly reorganizing pixels according to semantic similarity.

Recent approaches have begun to explore more compact representations. DSCC groups spectrally similar and spatially adjacent pixels into spectral supertokens and performs classification using a Transformer--Mamba architecture~\cite{liu2026hyperspectral}. However, assigning one supertoken prediction to all constituent pixels can suppress local variations around semantic boundaries. WIMNet directly processes complete HSI and LiDAR images but still flattens regularly arranged feature cells into dense Mamba sequences rather than constructing content-adaptive semantic tokens~\cite{li2026wimnet}. Similarly, MambaHSI+ organizes pixels or regular feature cells into row- and column-wise spatial sequences and models spectral bands bidirectionally~\cite{wang2025mambahsiplus}. Therefore, although patch-free learning improves efficiency and global context modeling, most existing methods still rely on dense or regularly structured token representations.

\subsection{Token Definition in Mamba Model} \label{token define}

Token definition and permutation are particularly important for SSMs because multidimensional visual features must be transformed into 1D sequences before state propagation. Existing studies mainly improve this process through directional scanning, spatial--spectral decomposition, multiscale serialization, or adaptive tokenization.

To overcome standard raster scanning, RSDehamba constructs multiple directional sequences from different image corners~\cite{zhou2024rsdehamba}, while SS-Mamba separately defines spatial and spectral token sequences to preserve complementary dependencies in hyperspectral cubes~\cite{huang2024spectral}. P-Mamba serializes multiscale feature maps into a unified sequence with scale embeddings~\cite{ma2025remote}. Multi-scale superpixel Mamba models further exploit hierarchical region structures~\cite{shi2026hsmmamba}, whereas supertoken-based Mamba--Transformer approaches progressively cluster local features into compact representations to reduce token redundancy~\cite{liu2026hyperspectral}. However, representing an entire clustered region with a single supertoken may discard fine-scale boundary information that is important for dense classification.

Other approaches construct higher-level or adaptive tokens. Semantic Tokenization-Based Mamba aggregates local patch information into semantic token sequences using center-pixel-guided scanning~\cite{ming2025semantic}, while RoMA introduces rotation-aware tokens through adaptive patch cropping and angle-aware embeddings~\cite{wang2025roma}. Dynamic Token Augmentation Mamba selectively retains object-informative tokens to improve robustness~\cite{huang2024dynamic}. Graph-based tokenization has also been explored to encode spatial--spectral relationships before linearizing graph features for Mamba modeling~\cite{ahmad2025hybrid}.

Despite these advances, most tokenization strategies remain strongly constrained by geometric locality or predefined reference pixels. In remote-sensing imagery, however, neighboring pixels may belong to different semantic classes because of spatial heterogeneity and geospatial non-stationarity~\cite{brunsdon1996geographically,estoque2014geospatial,zhang2026snstfm}. Consequently, predefined spatial sequences can introduce abrupt feature transitions into state propagation. Moreover, the contribution of historical information may decay as sequence distance increases~\cite{tu2026spectral,ali2025hidden}, making the construction of meaningful token relationships particularly important for Mamba models~\cite{liu2025defmamba,wu2026scan,li2026damamba}.

As illustrated in \autoref{Idea_fig}, conventional vision Mamba models typically construct long sequences according to spatial organization, potentially mixing semantically heterogeneous regions. In contrast, our method first clusters deep features into coherent semantic groups and then dynamically selects sparse tokens from each group. Each selected token therefore represents a semantic concept rather than a fixed grid cell, enabling Mamba to model long-range dependencies within more homogeneous sequences.

\section{Methodology} \label{method_part}

\begin{figure*}[]
    \centering
    \includegraphics[width=\textwidth]{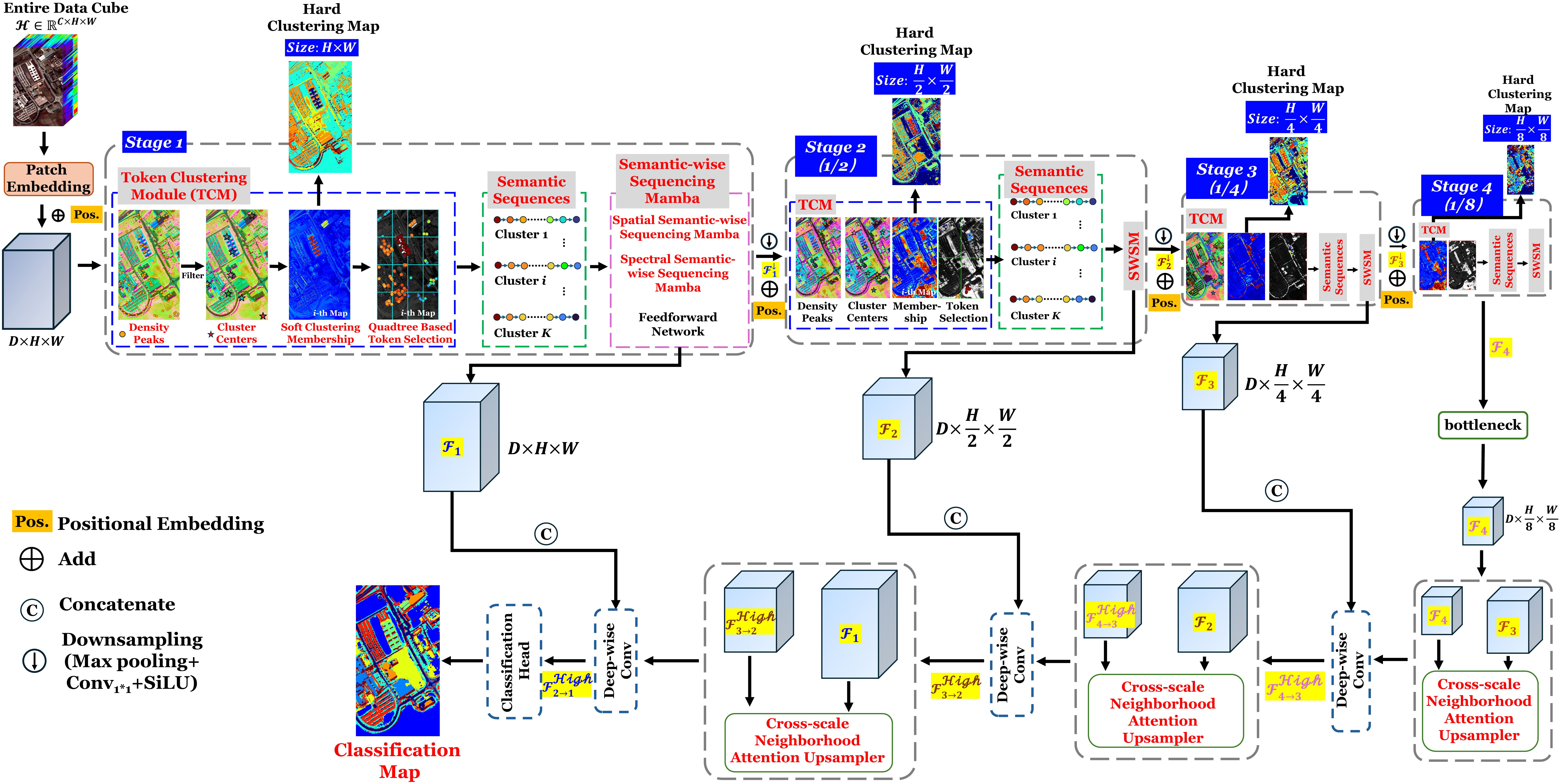}
    \caption{Overview of the proposed STMamba architecture. It consists of four hierarchical stages with Token Clustering Modules (TCM) for semantic token grouping, quadtree-based sparse token selection, and Spatial and Spectral Semantic-wise Sequencing Mamba modules (Spa-SWSM and Spe-SWSM) for spatial--spectral modeling. Hierarchical features are progressively encoded and subsequently recovered by a cross-scale Neighborhood Attention upsampler to generate the final prediction map.} 
    \label{Model Overview}
\end{figure*}

\subsection{Preliminaries}

State-space models (SSMs) provide an efficient framework for modeling long-range dependencies in sequential data. Given an input sequence $\mathbf{x}=(x_1,\ldots,x_L)$, the discretized SSM can be written as
\begin{align}
\begin{aligned}
\mathbf{h}_t
&=
\overline{\mathbf{A}}\mathbf{h}_{t-1}
+ \overline{\mathbf{B}}x_t, \
y_t =
\mathbf{C}\mathbf{h}_t
\end{aligned}
\label{eq:ssm_recurrence}
\end{align}
where $\overline{\mathbf{A}}$, $\overline{\mathbf{B}}$, and $\mathbf{C}$ denote the state-transition, input projection, and readout projection matrices, respectively.

Mamba \cite{gu2024mamba} further introduces the Selective State-Space (S6) mechanism, in which the state-space parameters become input-dependent and vary across sequence positions:
\begin{align}
\begin{aligned}
\mathbf{h}_t
&=
\overline{\mathbf{A}}_t\mathbf{h}_{t-1}
+ \overline{\mathbf{B}}_t x_t, \
y_t =
\mathbf{C}_t\mathbf{h}_t
\end{aligned}
\label{eq:s6_recurrence}
\end{align}

By unrolling \autoref{eq:s6_recurrence}, the selective state-space
operation can be equivalently represented in matrix form as
\begin{align}
    \mathbf{y}
    =
    \widetilde{\boldsymbol{\alpha}}\mathbf{x},
    \qquad
    \widetilde{\boldsymbol{\alpha}}
    \in\mathbb{R}^{L\times L}
    \label{eq:s6_matrix}
\end{align}
where $\widetilde{\boldsymbol{\alpha}}$ is defined by the following matrix:
\begin{align}
\widetilde{\boldsymbol{\alpha}}
=
\begin{bmatrix}
\mathbf{C}_1\overline{\mathbf{B}}_1
&
0
&
\cdots
&
0
\\
\mathbf{C}_2\overline{\mathbf{A}}_2\overline{\mathbf{B}}_1
&
\mathbf{C}_2\overline{\mathbf{B}}_2
&
\cdots
&
0
\\
\vdots
&
\vdots
&
\ddots
&
0
\\
\mathbf{C}_L
\displaystyle\prod_{k=2}^{L}
\overline{\mathbf{A}}_k
\overline{\mathbf{B}}_1
&
\mathbf{C}_L
\displaystyle\prod_{k=3}^{L}
\overline{\mathbf{A}}_k
\overline{\mathbf{B}}_2
&
\cdots
&
\mathbf{C}_L\overline{\mathbf{B}}_L
\end{bmatrix}
\label{eq:s6_matrix_expanded}
\end{align}

Each element of $\widetilde{\boldsymbol{\alpha}}$ characterizes the influence of an input token on an output token \cite{ali2025hidden}. Specifically,
\begin{align}
    \widetilde{\alpha}_{i,j}
    =
    \mathbf{C}_i
    \left(
    \prod_{k=j+1}^{i}
    \overline{\mathbf{A}}_k
    \right)
    \overline{\mathbf{B}}_j,
    \qquad j\leq i,
    \label{eq:s6_influence}
\end{align}
where $\widetilde{\alpha}_{i,j}$ represents the influence of the $j$-th input token $x_j$ on the $i$-th output token $y_i$.

Since $\overline{\mathbf{A}}_k$ is obtained through discretization, e.g.,
$\overline{\mathbf{A}}_k=\exp(\Delta_k\mathbf{A})$, and $\mathbf{A}$ typically has negative decay parameters \cite{dao2024transformers}, the product \(\prod_{k=j+1}^{i}\overline{\mathbf{A}}_k\)
can progressively decrease as the sequence distance $i-j$ increases. Consequently, the influence of distant tokens may decay along a long sequence. This motivates constructing shorter and semantically coherent token sequences, and the corresponding information-decay behavior is further analyzed in \autoref{decay}.

\subsection{Overview Architecture} 
STMamba adopts a four-stage hierarchical encoder-decoder architecture, as illustrated in \autoref{Model Overview}.
The input data is the entire hyperspectral image cube \(\mathcal{H} \in \mathbb{R}^{C \times H \times W}\), \(C, H, W\) represent the number of spectral channels, height, and width, respectively. STMamba consists of 4 hierarchical stages, an attention-based bottleneck,
and a Cross-scale Neighborhood Attention Upsampler for Segmentation (CNA Upsampler). Each stage contains two parallel Spatial/Spectral Semantic-wise Sequencing Mamba (Spa-SWSM, Spe-SWSM) and a feed-forward layer, and generates 4 multi-scale features \(\mathcal{F}_1, \mathcal{F}_2, \mathcal{F}_3, \mathcal{F}_4\). At each stage, TCM constructs semantic sequences that are subsequently processed by SWSM. Between two adjacent stages, a downsampling operation is used to construct hierarchical features and progressive clustering. The self-attention-based bottleneck is used to model the global context on the lowest-resolution feature \(\mathcal{F}_4\). The decoder recovers the low-resolution feature under the guidance of the previous stage's feature map by using a parameter-free CNA upsampler, and finally outputs the prediction logits.

\subsection{Token Clustering Module (TCM)} \label{TCM}

The Token Clustering Module (TCM) consists of two processes: clustering center determination using density peak clustering \cite{zeng2022not} and soft clustering membership generation. As shown in \autoref{Idea_fig}(b), representative semantic centers are first identified from sampled anchors, and soft memberships are then obtained by measuring the cosine similarity between the centers and all available tokens.

\subsubsection{Clustering Center Determination}

We employ k-nearest-neighbor density peaks clustering (DPC-KNN) to organize tokens into semantic groups. Directly applying DPC-KNN to all tokens requires approximately $\mathcal{O}(N^2D)$ complexity, where $N$ and $D$ denote the number of valid tokens and feature dimension, respectively. To reduce this cost, sparse grid-based sampling with ratio $\lambda_g$ is first applied to $\mathcal{F}_i$. A random spatial offset is introduced in each sampling round to alleviate deterministic grid-sampling bias.

Let
\begin{align}
    \mathcal{T}_i^p =
    \left\{
    \mathbf{x}_{i,1}^{p},
    \mathbf{x}_{i,2}^{p},
    \ldots,
    \mathbf{x}_{i,N_g}^{p}
    \right\}
\end{align}
denote the anchor set obtained from the $p$-th sampling at stage $i$, where $N_g \approx \lambda_g N$. The local density of each sampled anchor is estimated as
\begin{align}
    \rho_m^{p}
    =
    \exp\left(
    -\frac{1}{k}
    \sum_{\mathbf{x}_{i,n}^{p}
    \in
    \mathrm{KNN}(\mathbf{x}_{i,m}^{p})}
    d\left(
    \mathbf{x}_{i,m}^{p},
    \mathbf{x}_{i,n}^{p}
    \right)^2
    \right),
\end{align}
where $d(\cdot,\cdot)$ denotes the feature-space distance. Its minimum distance to an anchor with higher density is
\begin{align}
    \delta_m^{p}
    =
    \min_{\rho_j^{p}>\rho_m^{p}}
    d\left(
    \mathbf{x}_{i,m}^{p},
    \mathbf{x}_{i,j}^{p}
    \right).
\end{align}
The density-peaks score is then
\begin{align}
\gamma_m^{p}
=
\rho_m^{p}\delta_m^{p}
\end{align}
and the $K$ anchors with the largest scores are selected as candidate centers:
\begin{align}
\mathbf{M}^{p}
=
\left[
\boldsymbol{\mu}_1^{p},
\boldsymbol{\mu}_2^{p},
\ldots,
\boldsymbol{\mu}_K^{p}
\right]
\in
\mathbb{R}^{K\times D}
\end{align}

To reduce sensitivity to a specific sampling pattern, the procedure is repeated for $P$ independently offset grid samplings. The candidate centers are aggregated as
\begin{align}
\mathbf{M}_{\mathrm{agg}}
=
\left[
\mathbf{M}^{1};
\mathbf{M}^{2};
\ldots;
\mathbf{M}^{P}
\right]
\in
\mathbb{R}^{PK\times D}
\end{align}
and a second DPC-KNN operation is applied to obtain the final $K$ centers:
\begin{align}
\mathbf{M}
=
\left[
\boldsymbol{\mu}_1,
\boldsymbol{\mu}_2,
\ldots,
\boldsymbol{\mu}_K
\right]
\in
\mathbb{R}^{K\times D}
\end{align}

The resulting clustering complexity is approximately
\begin{align}
\mathcal{O}
\left(
P(\lambda_g N)^2D
+
(PK)^2D
\right),
\end{align}
which is substantially lower than directly clustering all $N$ tokens. In our experiments, the number of clusters $K$ is set to the number of semantic classes as a unified semantic-capacity setting and does not assume a one-to-one correspondence between clusters and classes.

\subsubsection{Soft Clustering Membership}

The soft clustering membership is obtained using cosine similarity:
\begin{align}
    \begin{aligned}
        \boldsymbol{s}_{ik} = \frac{\boldsymbol{f}_i^T \boldsymbol{\mu}_k}{||\boldsymbol{f}_i||_2 ||\boldsymbol{\mu}_k||_2}, i=1,2,3,4, k=1,..,K
    \end{aligned}
\end{align}
The corresponding soft semantic membership is calculated as
\begin{align}
    \begin{aligned}
        p_{ik} = \frac{\text{exp}(\boldsymbol{s}_{ik})}{\sum_{j=1}^{K} \text{exp}(\boldsymbol{s}_{ij})}
    \end{aligned}
\end{align}
The resulting membership $\boldsymbol{P} \in \mathbb{R}^{N_i \times K}$ provides dense token-to-cluster relationships for subsequent semantic sequence modeling.
\subsubsection{Clustering Center Diversity Loss}

To improve the discriminability of the cluster centers $\boldsymbol{\mu}\in\mathbb{R}^{K\times D}$, we introduce a center diversity loss based on the average squared cosine similarity between distinct centers. At the $i$-th stage, it is defined as
\begin{align}
\mathcal{L}_{\mathrm{div}}^{i}
=
\frac{1}{K(K-1)}
\sum_{m=1}^{K}
\sum_{\substack{n=1\\ n\neq m}}^{K}
\left(
\hat{\boldsymbol{\mu}}_m^{\top}
\hat{\boldsymbol{\mu}}_n
\right)^2,
\quad i=1,2,3,4.
\end{align}
Minimizing $\mathcal{L}_{\mathrm{div}}^{i}$ encourages different centers to occupy distinct directions in the embedding space. Since STMamba contains four stages, the overall diversity loss is
\begin{align}
    \begin{aligned}
        \mathcal{L}_{\text{div}} = \frac{1}{4} \sum_{i} \mathcal{L}_{\text{div}}^{i}, i=1,2,3,4
    \end{aligned}
\end{align}
Total loss is \(\mathcal{L}_{\mathrm{total}}=
    \mathcal{L}_{\mathrm{CE}}
    + \mathcal{L}_{\text{div}}\),
where, \(\mathcal{L}_{\mathrm{CE}}\) is the cross-entropy loss.

\subsection{Quadtree-based Adaptive Sparse Token Selection} \label{quadtree}

To construct sparse but spatially representative semantic sequences, we introduce a quadtree-based adaptive token selection strategy. For each semantic cluster, its soft membership map is used as a spatial score map indicating how strongly each token belongs to that cluster.

Instead of globally selecting only the highest-scoring tokens, which may concentrate samples within a small spatial region \cite{tang2022quadtree}, the membership map is recursively partitioned using a quadtree. Starting from the whole feature map, each region is divided into four sub-regions until the predefined minimum window size \(W_{\text{quad-tree}}\) or maximum tree depth \(D_{\mathrm{quad-tree}}\) is reached. Representative high-membership tokens are then selected from different spatial regions.

Given a global sparse ratio \(\lambda_s\), the number of selected tokens for the \(k\)-th semantic cluster is
\begin{align}
\begin{aligned}
M_k = \left\lfloor
\frac{N_i\ \times \lambda_s}{K}
\right\rfloor, k=1,..., K.
\end{aligned}
\label{sparse_ratio}
\end{align}
Thus, \(\lambda_s\) directly controls the overall token sparsity. The selected tokens preserve both semantic relevance and spatial diversity, and are subsequently ordered according to their membership scores and fed into the SWSM modules.

\subsection{Positional Embedding}

To preserve the spatial information of dynamically selected tokens, positional encoding is added to the intermediate feature map $\boldsymbol{F}\in\mathbb{R}^{H\times W\times D}$. The horizontal and vertical coordinates are normalized to $[-1,1]$, and each location $(i,j)$ is represented by
\begin{align}
\begin{aligned}
\left[
x_{i,j},
y_{i,j},
\sin(\pi x_{i,j}),
\cos(\pi x_{i,j}),
\sin(\pi y_{i,j}),
\cos(\pi y_{i,j})
\right]
\end{aligned}
\end{align}

The positional descriptor is projected to the feature dimension using $\mathrm{Conv}*{1\times1}$, producing $\mathbf{Pos}\in\mathbb{R}^{H\times W\times D}$. The spatially enhanced feature is then
\begin{equation}
\boldsymbol{F}_{\mathrm{pos}}
=
\boldsymbol{F}
+
\alpha \mathbf{Pos}
\end{equation}
where $\alpha$ is a learnable scaling parameter initialized with a small value to avoid strongly perturbing the original features at the beginning of training.

\subsection{Semantic-wise Sequencing Mamba (SWSM)} \label{SWSM}

The quadtree-based strategy in \autoref{quadtree} provides sparse tokens and their spatial indices for each semantic cluster. Within the \(k\)-th cluster, the selected tokens are sorted in descending order according to their clustering membership scores, forming a semantic sequence
\begin{align}
    \boldsymbol{X}_k^{s}
    =
    \left[
    \boldsymbol{x}_{i_1^{(k)}},
    \boldsymbol{x}_{i_2^{(k)}},
    \ldots,
    \boldsymbol{x}_{i_{M_k}^{(k)}}
    \right]^{\top}
    \in
    \mathbb{R}^{M_k\times D}
\end{align}
Unlike conventional spatial scanning, this ordering places tokens with similar semantic characteristics into the same sequence, thereby reducing abrupt feature transitions during state propagation.

To further exploit the confidence of each token belonging to its semantic cluster, the corresponding membership scores are transformed into a relative membership gate \(\boldsymbol{g}_k^{\mathrm{mem}}\). Instead of being used as an additional output gate, this semantic gate directly modulates the input and readout parameters of the selective state-space model:
\begin{align}
    \widetilde{\boldsymbol{B}}_{k,m}
    &=
    g_{k,m}^{\mathrm{mem}}
    \boldsymbol{B}_{k,m}
    \\
    \widetilde{\boldsymbol{C}}_{k,m}
    &=
    g_{k,m}^{\mathrm{mem}}
    \boldsymbol{C}_{k,m}
\end{align}
Therefore, tokens that are more strongly associated with the corresponding semantic cluster contribute more to state evolution and readout, while weakly associated tokens are suppressed. The resulting membership-conditioned Mamba operation is denoted as
\begin{align}
\boldsymbol{Y}_k
=
\mathcal{M}
\left(
\boldsymbol{X}_k^{s},
\boldsymbol{g}_k^{\mathrm{mem}}
\right)
\end{align}

\subsubsection{Spa-SWSM}

Spatial Semantic-wise Sequencing Mamba (Spa-SWSM) models long-range spatial dependencies among sparse tokens belonging to the same semantic cluster. The \(K\) semantic sequences are processed independently by forward Mamba scanning after reshaping them into \(\mathbb{R}^{K\times M\times D}\). Because the sequence is constructed according to semantic membership rather than raster position, spatially distant but semantically similar tokens can directly interact within the same state-space sequence.

After scanning, the processed sparse tokens are mapped back to their original spatial locations using the indices generated by the quadtree selection:
\begin{align}
    \boldsymbol{F}_{i}^{\mathrm{spa}}
    =
    \operatorname{ScatterAdd}
    \left(
    \left\{
    \boldsymbol{Y}_{k}^{\mathrm{spa}},
    \mathcal{I}_k
    \right\}_{k=1}^{K}
    \right) + \boldsymbol{X}_{k}^{s}
\end{align}

\subsubsection{Spe-SWSM}

In parallel, Spectral Semantic-wise Sequencing Mamba (Spe-SWSM) models complementary dependencies along the spectral-feature dimension. The same semantic tokens are reorganized into
\(\mathbb{R}^{KM\times S_{\mathrm{spe}}\times D_g}\),
where
\(S_{\mathrm{spe}}D_g=D\).
To capture cross-band information at different scales, \(D_g\) is set to \(2,4,4,8\) for the four stages, respectively. The spectral outputs are subsequently scattered back to their original token locations in the same manner as Spa-SWSM.

Through the parallel Spa-SWSM and Spe-SWSM branches, STMamba captures complementary spatial and spectral dependencies within semantically coherent sparse sequences. Their outputs are fused with the stage feature \(\mathcal{F}_i\) and subsequently processed by an FFN. At the lowest-resolution stage, a multi-head self-attention bottleneck is further applied to \(\mathcal{F}_4\) for global context modeling.

\subsection{Cross-scale Neighborhood Attention (CNA) Upsampler}

Since SWSM processes only a small fraction of spatial tokens (\(\lambda_s=0.01\)), dense spatial details must be progressively recovered in the decoder. We therefore adopt the parameter-free cross-scale neighborhood attention filtering operation, which uses the corresponding high resolution encoder feature as guidance to upsample the low resolution feature.

Taking the upsampling from \(\mathcal{F}_4 \in \mathbb{R}^{D \times \frac{H}{8} \times \frac{W}{8}}\) to the resolution of \(\mathcal{F}_3 \in \mathbb{R}^{D \times \frac{H}{4} \times \frac{W}{4}}\) as an example, the reconstructed feature \(\mathcal{F}^{\mathrm{High}}_{4\rightarrow3}\) is computed by neighborhood attention:
\begin{align}
    \begin{aligned}
        &\mathcal{F}^{\mathrm{High}}_{4\rightarrow3, p} = \frac{1}{Z(p)} \sum_{q \in \mathcal{N}(p)} \exp (\frac{<\mathcal{Q}_p, \mathcal{K}_q>}{\sqrt{D}}) \mathcal{F}_4
    \end{aligned}
\end{align}
where \(\mathcal{Q}_p=\mathcal{F}_{3,p}\), while \(\mathcal{K}_q\) is obtained by average pooling \(\mathcal{F}_3\) onto the low-resolution grid. Thus, the high-resolution encoder feature provides the queries and keys, whereas \(\mathcal{F}_4\) provides the attention values. \(Z(p)\) denotes the normalization factor and \(\mathcal{N}(p)\) is a local neighborhood.

The CNA operation is applied progressively across decoder stages to recover dense spatial features, followed by a linear layer for pixel-wise classification. Qualitative results are shown in \autoref{fig:upsampler_hu13}, with additional comparison against bilinear interpolation.




\label{Experiments}

\begin{table}[]
\centering
\caption{Land-cover classes and sample numbers of the three datasets.}
\label{data_info2}

\setlength{\tabcolsep}{2pt}
\renewcommand{\arraystretch}{0.92}
\scriptsize

\resizebox{\columnwidth}{!}{
\begin{tabular}{c l c c | c l c c}
\toprule

\multicolumn{8}{c}{\textbf{PU}} \\
\midrule
\textbf{No.} & \textbf{Class} & \textbf{Color} & \textbf{Tr./Val./Te.} &
\textbf{No.} & \textbf{Class} & \textbf{Color} & \textbf{Tr./Val./Te.} \\
\midrule

1 & Asphalt              & \cellcolor{pav1} & 5/5/6621
& 6 & Bare soil            & \cellcolor{pav6} & 5/5/5019 \\

2 & Meadows              & \cellcolor{pav2} & 5/5/18639
& 7 & Bitumen              & \cellcolor{pav7} & 5/5/1320 \\

3 & Gravel               & \cellcolor{pav3} & 5/5/2089
& 8 & Self-blocking bricks & \cellcolor{pav8} & 5/5/3672 \\

4 & Trees                & \cellcolor{pav4} & 5/5/3054
& 9 & Shadows              & \cellcolor{pav9} & 5/5/937 \\

5 & Painted metal sheets & \cellcolor{pav5} & 5/5/1335
&   &                      &                  & \\

\cmidrule(lr){1-8}
\multicolumn{6}{r}{\textbf{Total}} &
\multicolumn{2}{c}{\textbf{45/45/42686}} \\

\midrule
\multicolumn{8}{c}{\textbf{HU13}} \\
\midrule

1 & Healthy grass   & \cellcolor{h13c01} & 30/5/1216
& 9 & Road           & \cellcolor{h13c09} & 30/5/1217 \\

2 & Stressed grass  & \cellcolor{h13c02} & 30/5/1219
& 10 & Highway       & \cellcolor{h13c10} & 30/5/1192 \\

3 & Synthetic grass & \cellcolor{h13c03} & 30/5/672
& 11 & Railway       & \cellcolor{h13c11} & 30/5/1200 \\

4 & Trees           & \cellcolor{h13c04} & 30/5/1209
& 12 & Parking Lot 1 & \cellcolor{h13c12} & 30/5/1198 \\

5 & Soil            & \cellcolor{h13c05} & 30/5/1207
& 13 & Parking Lot 2 & \cellcolor{h13c13} & 30/5/434 \\

6 & Water           & \cellcolor{h13c06} & 30/5/290
& 14 & Tennis court  & \cellcolor{h13c14} & 30/5/393 \\

7 & Residential     & \cellcolor{h13c07} & 30/5/1233
& 15 & Running track & \cellcolor{h13c15} & 30/5/625 \\

8 & Commercial      & \cellcolor{h13c08} & 30/5/1209
&    &               &                  & \\

\cmidrule(lr){1-8}
\multicolumn{6}{r}{\textbf{Total}} &
\multicolumn{2}{c}{\textbf{450/75/14514}} \\

\midrule
\multicolumn{8}{c}{\textbf{HU18}} \\
\midrule

1 & Healthy grass
& \cellcolor{h18c01} & 30/5/9764
& 11 & Sidewalks
& \cellcolor{h18c11} & 30/5/33967 \\

2 & Stressed grass
& \cellcolor{h18c02} & 30/5/32467
& 12 & Crosswalks
& \cellcolor{h18c12} & 30/5/1481 \\

3 & Artificial turf
& \cellcolor{h18c03} & 30/5/649
& 13 & Major thoroughfares
& \cellcolor{h18c13} & 30/5/46323 \\

4 & Evergreen trees
& \cellcolor{h18c04} & 30/5/13553
& 14 & Highways
& \cellcolor{h18c14} & 30/5/9814 \\

5 & Deciduous trees
& \cellcolor{h18c05} & 30/5/5013
& 15 & Railways
& \cellcolor{h18c15} & 30/5/6902 \\

6 & Bare earth
& \cellcolor{h18c06} & 30/5/4481
& 16 & Paved parking lots
& \cellcolor{h18c16} & 30/5/11440 \\

7 & Water
& \cellcolor{h18c07} & 30/5/231
& 17 & Unpaved parking lots
& \cellcolor{h18c17} & 30/5/114 \\

8 & Residential buildings
& \cellcolor{h18c08} & 30/5/39727
& 18 & Cars
& \cellcolor{h18c18} & 30/5/6543 \\

9 & Non-residential buildings
& \cellcolor{h18c09} & 30/5/223649
& 19 & Trains
& \cellcolor{h18c19} & 30/5/5330 \\

10 & Roads
& \cellcolor{h18c10} & 30/5/45775
& 20 & Stadium seats
& \cellcolor{h18c20} & 30/5/6789 \\

\cmidrule(lr){1-8}
\multicolumn{6}{r}{\textbf{Total}} &
\multicolumn{2}{c}{\textbf{600/100/504012}} \\

\bottomrule
\end{tabular}
}

\vspace{1mm}
\begin{minipage}{\columnwidth}
\scriptsize
\textit{Note:} Tr., Val., and Te. denote the numbers of training,
validation, and testing samples, respectively.
\end{minipage}

\end{table}

\section{Experimental results and analysis} \label{exp}
\subsection{Datasets and Evaluation Metrics}
To thoroughly assess the effectiveness of the proposed approach, we conduct experiments on three large-scale benchmark hyperspectral datasets, namely Pavia University (PU), Houston2013 (HU13) \cite{debes2014hyperspectral}, and Houston2018 (HU18) \cite{le20182018}. The training samples of each dataset are shown and listed in \autoref{data_info2}. In the following experiments, 5 samples per class of the PU dataset are selected, and 30 samples per class of the HU13 and HU18 are selected, respectively. The number of validation samples is set to 5, and the rest are for testing, as listed in \autoref{data_info2}.

\subsubsection{Pavia University} It was acquired by the ROSIS sensor over PU and its surroundings, Pavia, Italy. This dataset has 103 spectral bands ranging from 430 to 860 nm. Its spatial resolution is 1.3 m, and its image size is 610 \(\times\) 340. Nine land-cover categories are covered.


\subsubsection{Houston2013} The HU13 dataset was captured over the University of Houston and its neighboring areas using the ITRES CASI 1500 hyperspectral imager. The image comprises 144 spectral bands with a spatial dimension of 349 \(\times\) 1905. This dataset was officially released as part of the 2013 IEEE Geoscience and Remote Sensing Society (GRSS) Data Fusion Contest and contains 15 land-cover categories. 

\subsubsection{Houston2018} The HU18 dataset was provided by the 2018 IEEE GRSS Data Fusion Contest and acquired by the National Center for Airborne Laser Mapping over the University of Houston campus and its neighborhood. The HSI covers a 380–1050 nm spectral wavelength range with 48 bands at a 1 m ground sampling distance of size 601 \(\times\) 2384. The dataset contains 20 land-cover classes.
To quantitatively evaluate the proposed method and other compared methods, we choose the following commonly used metrics, i.e., Overall Classification Accuracy (OA \(\uparrow\)), Average Classification Accuracy (AA \(\uparrow\)), per-class accuracy, and Kappa Coefficient (\(\kappa\) \(\uparrow\)).

\begin{table*}[]
\centering
\caption{Quantitative classification results on the PU dataset.}
\label{tab:pavia_results}

\renewcommand{\arraystretch}{1.12}
\setlength{\tabcolsep}{2.4pt}

\resizebox{\textwidth}{!}{
\begin{tabular}{ccc ccc cccccc cc c}
\toprule

\multirow{2}{*}{\textbf{Class}}
&
\multirow{2}{*}{\textbf{Color}}
&
\multirow{2}{*}{\makecell{\textbf{Train}\\\textbf{Num.}}}
&
\multicolumn{1}{c}{\textbf{Traditional}}
&
\multicolumn{1}{c}{\textbf{Transformer}}
&
\multicolumn{1}{c}{\textbf{CNN}}
&
\multicolumn{6}{c}{\textbf{Mamba-based}}
&
\multicolumn{2}{c}{\textbf{Clustering-based}}
&
\multicolumn{1}{c}{\makecell{\textbf{Mamba +}\\\textbf{Clustering}}}
\\

\cmidrule(lr){4-4}
\cmidrule(lr){5-5}
\cmidrule(lr){6-6}
\cmidrule(lr){7-12}
\cmidrule(lr){13-14}
\cmidrule(lr){15-15}

&
&
&
\makecell{Random\\Forest}
&
\makecell{CPFormer\\\cite{11382038}}
&
\makecell{S$^{2}$VNet\\\cite{han2025subpixel}}
&
\makecell{SDMamba\\\cite{xu2025sparse}}
&
\makecell{MambaLG\\\cite{pan2025multiscale}}
&
\makecell{MambaHSI\\\cite{10604894}}
&
\makecell{MambaHSI+\\\cite{wang2025mambahsiplus}}
&
\makecell{S$^{2}$Mamba\\\cite{s2mamba}}
&
\makecell{DSCC\\\cite{liu2026hyperspectral}}
&
\makecell{PSFormer\\\cite{10695122}}
&
\makecell{DMSGer\\\cite{9927317}}
&
\textbf{Ours}
\\

\midrule

1 & \cellcolor{pav1}\phantom{00} & 5
& 66.24±5.52
& 72.44±6.70
& 74.06±12.31
& 78.40±6.29
& 77.88±5.42
& 77.35±13.27
& 81.18±11.34
& 76.20±7.29
& 69.66±8.59
& 74.85±8.81
& 55.81±\textbf{4.09}
& \cellcolor[RGB]{248,226,214}80.49±6.36
\\

2 & \cellcolor{pav2}\phantom{00} & 5
& 51.31±14.13
& 74.04±16.19
& 60.40±16.92
& 64.91±14.65
& 31.18±9.20
& 76.61±10.62
& 71.56±13.94
& 61.69±15.09
& 68.96±8.97
& 62.70±17.64
& \cellcolor[RGB]{248,226,214}84.39±\textbf{4.76}
& 78.86±10.04
\\

3 & \cellcolor{pav3}\phantom{00} & 5
& 48.29±11.90
& 74.51±11.23
& 84.30±11.58
& 73.45±12.64
& 92.62±2.17
& 66.57±14.81
& 77.20±12.81
& 75.25±12.42
& 80.64±10.54
& 89.23±13.45
& \cellcolor[RGB]{248,226,214}91.52±\textbf{6.25}
& 89.82±13.04
\\

4 & \cellcolor{pav4}\phantom{00} & 5
& 81.81±7.67
& 84.32±6.61
& 85.65±14.14
& 87.91±6.50
& 96.59±0.90
& 72.32±7.78
& 88.91±5.65
& 94.55±3.40
& 68.95±6.84
& 96.47±\textbf{2.61}
& 68.76±6.82
& \cellcolor[RGB]{248,226,214}96.61±2.89
\\

5 & \cellcolor{pav5}\phantom{00} & 5
& 98.96±0.39
& 99.56±0.82
& 99.23±0.98
& 98.76±0.98
& 99.87±0.27
& 99.67±0.51
& 98.50±2.88
& 99.70±0.80
& 96.04±4.03
& \cellcolor[RGB]{248,226,214}100.00±\textbf{0.00}
& 99.38±0.24
& 99.89±0.15
\\

6 & \cellcolor{pav6}\phantom{00} & 5
& 55.84±9.33
& 89.47±10.00
& 69.88±16.67
& 86.12±8.51
& 65.67±6.98
& 91.29±6.03
& 93.18±5.59
& 80.77±14.02
& 95.74±2.77
& 63.38±22.47
& 95.36±\textbf{1.89}
& \cellcolor[RGB]{248,226,214}99.18±1.95
\\

7 & \cellcolor{pav7}\phantom{00} & 5
& 85.06±8.14
& 99.15±1.46
& 94.40±5.10
& 89.39±6.95
& 95.54±1.65
& 83.51±10.88
& 91.02±8.47
& 95.43±3.81
& 95.08±5.82
& 88.88±14.77
& 90.52±8.38
& \cellcolor[RGB]{248,226,214}99.27±\textbf{1.29}
\\

8 & \cellcolor{pav8}\phantom{00} & 5
& 64.23±10.22
& 70.34±11.58
& 65.75±19.87
& 74.70±10.94
& 66.11±5.24
& 81.06±10.56
& 89.43±9.48
& 84.62±9.99
& 83.00±14.20
& 77.04±19.37
& 49.98±6.23
& \cellcolor[RGB]{248,226,214}95.11±\textbf{4.26}
\\

9 & \cellcolor{pav9}\phantom{00} & 5
& \cellcolor[RGB]{248,226,214}99.96±0.06
& 98.71±2.00
& 89.71±4.77
& 94.15±6.60
& 90.71±2.39
& 97.00±3.89
& 95.28±5.76
& 98.94±0.96
& 74.38±17.86
& 99.74±0.27
& 94.77±3.12
& 99.94±\textbf{0.05}
\\

\midrule

\multicolumn{3}{c}{\textbf{OA (\%)}}
& 60.91±5.20
& 77.96±6.80
& 70.18±7.50
& 74.86±6.81
& 58.81±4.24
& 79.42±5.17
& 80.61±5.81
& 74.26±6.20
& 75.77±3.60
& 72.42±5.72
& 78.64±\textbf{2.43}
& \cellcolor[RGB]{248,226,214}86.49±3.77
\\

\multicolumn{3}{c}{\textbf{AA (\%)}}
& 72.42±\textbf{1.28}
& 83.73±1.87
& 81.38±4.71
& 83.08±3.34
& 79.97±1.29
& 82.82±2.95
& 87.36±2.53
& 85.36±2.04
& 81.38±3.07
& 83.59±3.21
& 81.39±1.47
& \cellcolor[RGB]{248,226,214}93.26±1.64
\\

\multicolumn{3}{c}{\textbf{$\kappa \times 100$ (\%)}}
& 52.24±4.84
& 72.40±7.65
& 63.34±8.13
& 68.93±7.68
& 52.17±4.13
& 73.87±6.18
& 75.74±6.75
& 68.38±6.61
& 69.78±4.02
& 65.84±6.01
& 72.76±\textbf{2.83}
& \cellcolor[RGB]{248,226,214}82.91±4.45
\\

\bottomrule
\end{tabular}
}
\end{table*}

\begin{table*}[]
\centering
\caption{Quantitative classification results on the HU13 dataset.}
\label{tab:hu13_results}

\renewcommand{\arraystretch}{1.12}
\setlength{\tabcolsep}{2.4pt}

\resizebox{\textwidth}{!}{
\begin{tabular}{ccc ccc cccccc cc c}
\toprule

\multirow{2}{*}{\textbf{Class}}
&
\multirow{2}{*}{\textbf{Color}}
&
\multirow{2}{*}{\makecell{\textbf{Train}\\\textbf{Num.}}}
&
\multicolumn{1}{c}{\textbf{Traditional}}
&
\multicolumn{1}{c}{\textbf{Transformer}}
&
\multicolumn{1}{c}{\textbf{CNN}}
&
\multicolumn{6}{c}{\textbf{Mamba-based}}
&
\multicolumn{2}{c}{\textbf{Clustering-based}}
&
\multicolumn{1}{c}{\makecell{\textbf{Mamba +}\\\textbf{Clustering}}}
\\

\cmidrule(lr){4-4}
\cmidrule(lr){5-5}
\cmidrule(lr){6-6}
\cmidrule(lr){7-12}
\cmidrule(lr){13-14}
\cmidrule(lr){15-15}

&
&
&
\makecell{Random\\Forest}
&
\makecell{CPFormer\\\cite{11382038}}
&
\makecell{S$^{2}$VNet\\\cite{han2025subpixel}}
&
\makecell{SDMamba\\\cite{xu2025sparse}}
&
\makecell{MambaLG\\\cite{pan2025multiscale}}
&
\makecell{MambaHSI\\\cite{10604894}}
&
\makecell{MambaHSI+\\\cite{wang2025mambahsiplus}}
&
\makecell{S$^{2}$Mamba\\\cite{s2mamba}}
&
\makecell{DSCC\\\cite{liu2026hyperspectral}}
&
\makecell{PSFormer\\\cite{10695122}}
&
\makecell{DMSGer\\\cite{9927317}}
&
\textbf{Ours}
\\

\midrule

1 & \cellcolor{h13c01}\phantom{00} & 30
& 93.71±2.73
& 90.26±4.58
& 95.91±4.82
& 91.47±6.73
& 95.24±4.43
& 83.46±1.78
& 85.16±\textbf{1.57}
& 89.90±5.22
& 82.80±6.85
& 94.57±5.36
& \cellcolor[RGB]{248,226,214}96.26±1.58
& 94.32±5.31
\\

2 & \cellcolor{h13c02}\phantom{00} & 30
& 93.58±4.54
& 93.82±4.88
& 92.61±5.58
& 90.53±4.93
& 97.22±3.43
& 92.54±4.53
& 95.10±2.69
& 92.85±6.51
& 84.18±6.26
& 93.97±4.88
& \cellcolor[RGB]{248,226,214}97.72±\textbf{0.95}
& 97.19±1.73
\\

3 & \cellcolor{h13c03}\phantom{00} & 30
& 94.15±2.12
& 98.50±2.25
& 99.72±0.44
& 99.88±0.15
& 99.43±0.56
& 92.52±4.05
& 92.16±8.33
& 99.81±0.11
& 95.00±4.99
& \cellcolor[RGB]{248,226,214}100.00±\textbf{0.00}
& \cellcolor[RGB]{248,226,214}100.00±\textbf{0.00}
& 100.00±0.00
\\

4 & \cellcolor{h13c04}\phantom{00} & 30
& 94.24±2.41
& 94.29±4.33
& 93.44±2.05
& 93.42±2.44
& 96.57±2.78
& 96.91±2.08
& 96.66±2.29
& 93.51±\textbf{1.76}
& 81.15±5.34
& 97.71±2.27
& 89.53±2.56
& \cellcolor[RGB]{248,226,214}97.58±2.34
\\

5 & \cellcolor{h13c05}\phantom{00} & 30
& 91.95±4.73
& 99.06±0.84
& 99.32±1.29
& \cellcolor[RGB]{248,226,214}100.00±\textbf{0.00}
& 99.31±1.83
& 99.30±0.60
& 99.58±0.32
& \cellcolor[RGB]{248,226,214}100.00±\textbf{0.00}
& 97.18±3.29
& 99.61±1.16
& \cellcolor[RGB]{248,226,214}100.00±\textbf{0.00}
& 98.14±5.05
\\

6 & \cellcolor{h13c06}\phantom{00} & 30
& 93.65±2.88
& 94.11±6.08
& 90.93±4.31
& 87.10±2.14
& 98.41±1.21
& 97.55±2.11
& 97.21±3.48
& 89.31±4.09
& 95.83±3.94
& 98.57±1.23
& 99.41±\textbf{0.80}
& \cellcolor[RGB]{248,226,214}99.52±0.89
\\

7 & \cellcolor{h13c07}\phantom{00} & 30
& 75.82±4.68
& 85.48±6.30
& 74.17±6.67
& 87.46±1.50
& 93.36±2.96
& 88.85±3.26
& 93.40±3.44
& 91.41±3.56
& 89.06±4.29
& 86.20±6.22
& 93.97±\textbf{0.92}
& \cellcolor[RGB]{248,226,214}95.54±3.66
\\

8 & \cellcolor{h13c08}\phantom{00} & 30
& 63.23±6.36
& 78.87±5.15
& 72.48±3.93
& 56.69±5.63
& 82.82±4.23
& 77.76±3.07
& 81.79±5.38
& 61.11±7.40
& 79.77±8.18
& 83.17±\textbf{2.18}
& 73.11±3.08
& \cellcolor[RGB]{248,226,214}84.42±6.83
\\

9 & \cellcolor{h13c09}\phantom{00} & 30
& 72.59±4.22
& 76.76±9.85
& 79.35±6.69
& 81.84±4.95
& 90.90±2.87
& 87.78±5.39
& 93.43±4.14
& 82.79±3.36
& 77.13±5.83
& 89.64±5.21
& 81.94±\textbf{2.19}
& \cellcolor[RGB]{248,226,214}91.97±4.81
\\

10 & \cellcolor{h13c10}\phantom{00} & 30
& 73.16±5.83
& 90.52±6.33
& 89.70±7.98
& 51.93±17.79
& 96.14±2.71
& 73.30±16.13
& 88.66±5.37
& 69.57±12.19
& 98.80±1.41
& 95.27±3.02
& 98.19±1.13
& \cellcolor[RGB]{248,226,214}98.82±\textbf{1.02}
\\

11 & \cellcolor{h13c11}\phantom{00} & 30
& 73.35±7.03
& 91.45±4.52
& 73.49±14.44
& 72.32±7.71
& 94.08±2.94
& 65.56±1.30
& 75.42±7.08
& 82.40±9.47
& 94.57±3.48
& 92.63±4.11
& 90.88±\textbf{0.73}
& \cellcolor[RGB]{248,226,214}97.19±2.09
\\

12 & \cellcolor{h13c12}\phantom{00} & 30
& 57.58±5.97
& 90.96±3.93
& 78.87±7.49
& 65.29±12.32
& 91.54±4.08
& 86.18±3.89
& 88.44±6.51
& 57.85±11.27
& 92.99±3.98
& 88.52±8.48
& \cellcolor[RGB]{248,226,214}96.28±\textbf{1.23}
& 93.03±4.61
\\

13 & \cellcolor{h13c13}\phantom{00} & 30
& 41.93±4.40
& 83.12±5.46
& 88.47±6.13
& 95.62±\textbf{0.23}
& 97.13±1.60
& 93.82±1.60
& 95.16±2.58
& 92.41±2.18
& 94.70±3.66
& 86.90±3.19
& 93.78±1.27
& \cellcolor[RGB]{248,226,214}97.00±2.81
\\

14 & \cellcolor{h13c14}\phantom{00} & 30
& 96.46±1.73
& 99.63±0.52
& 99.92±0.11
& 99.77±0.31
& 99.90±0.23
& \cellcolor[RGB]{248,226,214}100.00±\textbf{0.00}
& \cellcolor[RGB]{248,226,214}100.00±\textbf{0.00}
& \cellcolor[RGB]{248,226,214}100.00±\textbf{0.00}
& 99.97±0.08
& \cellcolor[RGB]{248,226,214}100.00±\textbf{0.00}
& \cellcolor[RGB]{248,226,214}100.00±\textbf{0.00}
& \cellcolor[RGB]{248,226,214}100.00±\textbf{0.00}
\\

15 & \cellcolor{h13c15}\phantom{00} & 30
& 95.55±1.76
& 99.42±1.64
& 99.92±0.12
& \cellcolor[RGB]{248,226,214}100.00±\textbf{0.00}
& 99.70±0.43
& 97.50±2.30
& 96.24±2.55
& \cellcolor[RGB]{248,226,214}100.00±\textbf{0.00}
& 95.34±3.94
& \cellcolor[RGB]{248,226,214}100.00±\textbf{0.00}
& 97.92±1.06
& \cellcolor[RGB]{248,226,214}100.00±\textbf{0.00}
\\

\midrule

\multicolumn{3}{c}{\textbf{OA (\%)}}
& 80.03±1.48
& 90.20±2.44
& 86.89±1.82
& 82.32±2.59
& 94.60±1.05
& 86.89±1.31
& 90.76±1.10
& 84.71±1.49
& 89.09±1.19
& 93.00±0.99
& 92.86±0.52
& \cellcolor[RGB]{248,226,214}95.57±\textbf{0.71}
\\

\multicolumn{3}{c}{\textbf{AA (\%)}}
& 80.73±1.23
& 91.08±2.20
& 88.56±1.50
& 84.89±2.16
& 95.45±0.91
& 88.80±1.12
& 91.89±1.01
& 86.86±1.31
& 90.56±1.12
& 93.78±0.91
& 93.93±0.43
& \cellcolor[RGB]{248,226,214}96.31±\textbf{0.68}
\\

\multicolumn{3}{c}{\textbf{$\kappa \times 100$ (\%)}}
& 78.42±1.60
& 89.40±2.63
& 85.82±1.97
& 80.78±2.81
& 94.16±1.14
& 85.82±1.41
& 90.01±1.19
& 83.49±1.61
& 88.21±1.28
& 92.43±1.07
& 92.28±0.56
& \cellcolor[RGB]{248,226,214}95.21±\textbf{0.77}
\\

\bottomrule
\end{tabular}
}
\end{table*}

\begin{table*}[]
\centering
\caption{Quantitative classification results on the HU18 dataset.}
\label{tab:hu18_results}

\renewcommand{\arraystretch}{1.12}
\setlength{\tabcolsep}{2.4pt}

\resizebox{\textwidth}{!}{
\begin{tabular}{ccc ccc cccccc cc c}
\toprule

\multirow{2}{*}{\textbf{Class}}
&
\multirow{2}{*}{\textbf{Color}}
&
\multirow{2}{*}{\makecell{\textbf{Train}\\\textbf{Num.}}}
&
\multicolumn{1}{c}{\textbf{Traditional}}
&
\multicolumn{1}{c}{\textbf{Transformer}}
&
\multicolumn{1}{c}{\textbf{CNN}}
&
\multicolumn{6}{c}{\textbf{Mamba-based}}
&
\multicolumn{2}{c}{\textbf{Clustering-based}}
&
\multicolumn{1}{c}{\makecell{\textbf{Mamba +}\\\textbf{Clustering}}}
\\

\cmidrule(lr){4-4}
\cmidrule(lr){5-5}
\cmidrule(lr){6-6}
\cmidrule(lr){7-12}
\cmidrule(lr){13-14}
\cmidrule(lr){15-15}

&
&
&
\makecell{Random\\Forest}
&
\makecell{CPFormer\\\cite{11382038}}
&
\makecell{S$^{2}$VNet\\\cite{han2025subpixel}}
&
\makecell{SDMamba\\\cite{xu2025sparse}}
&
\makecell{MambaLG\\\cite{pan2025multiscale}}
&
\makecell{MambaHSI\\\cite{10604894}}
&
\makecell{MambaHSI+\\\cite{wang2025mambahsiplus}}
&
\makecell{S$^{2}$Mamba\\\cite{s2mamba}}
&
\makecell{DSCC\\\cite{liu2026hyperspectral}}
&
\makecell{PSFormer\\\cite{10695122}}
&
\makecell{DMSGer\\\cite{9927317}}
&
\textbf{Ours}
\\

\midrule

1  & \cellcolor{h18c01}\phantom{00} & 30
& 91.70±2.38
& 83.75±7.07
& 90.61±5.74
& \best{91.95±3.22}
& 89.82±3.18
& 82.31±5.77
& 82.34±6.12
& 91.37±2.92
& 72.70±6.26
& 90.29±4.46
& 89.72±\std{1.98}
& 91.14±2.52 \\

2  & \cellcolor{h18c02}\phantom{00} & 30
& \best{84.00±4.39}
& 71.97±6.20
& 81.09±8.07
& 82.06±3.71
& 79.98±\std{2.99}
& 79.80±6.95
& 81.74±4.62
& 75.13±5.48
& 71.08±5.27
& 71.55±24.26
& 60.96±4.01
& 81.38±4.29 \\

3  & \cellcolor{h18c03}\phantom{00} & 30
& 98.64±0.89
& 99.83±0.27
& 99.63±0.39
& 99.80±0.24
& 99.98±0.05
& 96.17±5.89
& 99.91±0.16
& 99.96±0.09
& 98.86±1.44
& 90.00±30.00
& \best{100.00±\std{0.00}}
& \best{100.00±\std{0.00}} \\

4  & \cellcolor{h18c04}\phantom{00} & 30
& 88.78±2.19
& 94.08±3.83
& 95.90±2.06
& 96.42±1.13
& 94.34±1.12
& 92.10±1.62
& \best{96.81±1.26}
& 96.36±\std{1.09}
& 85.59±4.09
& 82.00±27.47
& 87.85±2.33
& 94.51±1.88 \\

5  & \cellcolor{h18c05}\phantom{00} & 30
& 69.09±4.46
& 90.28±4.80
& 81.15±2.88
& 82.63±2.73
& 83.05±2.84
& 84.94±\std{1.39}
& 89.16±3.14
& 80.13±3.94
& 79.25±5.93
& 77.10±26.07
& 82.53±2.50
& \best{90.77±3.54} \\

6  & \cellcolor{h18c06}\phantom{00} & 30
& 81.39±3.21
& 97.62±3.00
& 93.12±5.72
& 96.09±3.88
& 96.53±2.59
& 78.47±12.81
& 87.72±6.73
& 95.60±2.33
& 97.49±2.64
& 88.50±29.59
& \best{99.74±\std{0.34}}
& 99.28±2.15 \\

7  & \cellcolor{h18c07}\phantom{00} & 30
& 96.53±1.77
& 99.76±0.75
& 99.52±0.68
& 99.13±1.11
& 99.91±0.18
& 99.21±0.92
& 99.44±0.76
& 99.87±0.38
& 99.78±0.52
& 89.66±29.89
& \best{100.00±\std{0.00}}
& 99.61±0.30 \\

8  & \cellcolor{h18c08}\phantom{00} & 30
& 65.04±5.47
& 74.26±7.88
& 69.90±5.63
& 77.93±3.50
& 78.52±\std{2.67}
& 57.79±9.45
& 77.41±3.24
& 80.98±4.26
& 90.86±3.26
& 73.94±24.80
& 82.73±3.10
& \best{92.38±2.89} \\

9  & \cellcolor{h18c09}\phantom{00} & 30
& 28.06±3.75
& \best{71.33±5.79}
& 53.23±6.24
& 56.50±6.37
& 65.55±5.51
& 45.13±5.84
& 57.21±2.81
& 36.11±6.81
& 61.88±4.89
& 57.80±20.43
& 27.91±\std{2.09}
& 68.05±5.12 \\

10 & \cellcolor{h18c10}\phantom{00} & 30
& 18.53±4.98
& 37.70±8.03
& 36.90±4.67
& 34.29±6.19
& 36.14±4.66
& 38.30±3.53
& \best{45.27±7.33}
& 21.14±9.58
& 34.33±4.57
& 38.90±13.82
& 26.94±\std{2.78}
& 43.91±5.85 \\

11 & \cellcolor{h18c11}\phantom{00} & 30
& 21.01±5.19
& 34.62±8.55
& 37.51±3.81
& 42.48±5.03
& 43.59±3.85
& 27.36±6.61
& 40.16±6.16
& 32.47±7.41
& 32.76±5.03
& 35.10±12.93
& 23.86±\std{1.71}
& \best{52.67±4.96} \\

12 & \cellcolor{h18c12}\phantom{00} & 30
& 35.38±4.33
& 54.76±8.60
& 45.22±11.25
& 48.85±4.20
& 62.83±4.24
& 56.04±3.23
& 64.63±11.65
& 50.70±5.44
& 67.92±5.82
& 58.93±20.22
& 67.84±\std{2.57}
& \best{76.36±4.96} \\

13 & \cellcolor{h18c13}\phantom{00} & 30
& 25.63±5.54
& 48.72±7.68
& 41.01±6.86
& 42.14±7.40
& 50.87±6.82
& 25.99±5.02
& 35.93±12.51
& 50.99±8.94
& 58.97±9.09
& 50.19±17.05
& 48.55±\std{3.71}
& \best{62.29±7.05} \\

14 & \cellcolor{h18c14}\phantom{00} & 30
& 65.55±6.56
& 88.78±10.37
& 81.68±6.80
& 90.60±4.79
& 91.63±5.04
& 71.29±13.53
& 89.41±4.96
& 93.36±5.74
& 95.28±4.22
& 81.51±27.45
& 95.30±\std{1.60}
& \best{96.46±3.01} \\

15 & \cellcolor{h18c15}\phantom{00} & 30
& 85.85±3.85
& 97.91±3.44
& 95.07±5.06
& 98.36±0.96
& 97.26±2.12
& 93.43±3.43
& 92.08±5.29
& 96.73±2.41
& 94.37±2.47
& 89.63±29.88
& 91.47±2.16
& \best{99.52±\std{0.74}} \\

16 & \cellcolor{h18c16}\phantom{00} & 30
& 58.93±5.22
& 83.77±11.89
& 82.82±5.32
& 85.38±\std{1.85}
& 90.17±1.99
& 74.77±11.30
& \best{92.62±2.73}
& 80.51±4.77
& 54.05±8.44
& 79.66±27.17
& 76.59±3.62
& 92.21±3.96 \\

17 & \cellcolor{h18c17}\phantom{00} & 30
& 95.26±1.76
& \best{100.00±\std{0.00}}
& 99.82±0.52
& \best{100.00±\std{0.00}}
& 99.53±1.06
& 99.71±0.65
& 95.99±5.38
& \best{100.00±\std{0.00}}
& 99.30±2.11
& 90.00±30.00
& \best{100.00±\std{0.00}}
& \best{100.00±\std{0.00}} \\

18 & \cellcolor{h18c18}\phantom{00} & 30
& 48.52±5.43
& 87.02±6.09
& 78.27±8.84
& 90.77±2.63
& \best{93.67±\std{1.87}}
& 85.84±2.63
& 90.93±2.33
& 88.19±3.78
& 83.16±10.33
& 80.66±27.06
& 82.75±2.97
& 87.21±2.93 \\

19 & \cellcolor{h18c19}\phantom{00} & 30
& 57.56±5.05
& 94.78±4.03
& 91.42±2.09
& 93.17±2.24
& 96.74±2.41
& 89.25±3.66
& 95.35±4.04
& 91.13±6.33
& 90.91±2.67
& 84.46±28.62
& 91.14±\std{1.26}
& \best{97.83±1.95} \\

20 & \cellcolor{h18c20}\phantom{00} & 30
& 80.60±5.72
& 98.24±1.63
& 92.48±7.20
& 93.70±2.13
& 99.50±0.34
& 91.06±5.46
& 94.66±4.35
& 95.83±1.22
& 98.45±2.03
& 88.97±29.76
& 99.60±0.27
& \best{100.00±\std{0.00}} \\

\midrule

\multicolumn{3}{c}{\textbf{OA (\%)}}
& 40.44±1.87
& 67.08±1.90
& 58.31±2.16
& 61.22±3.33
& 66.41±2.45
& 51.05±2.57
& 61.69±1.55
& 50.77±3.26
& 63.45±2.52
& 60.22±19.65
& 45.79±\std{1.25}
& \best{71.34±2.64} \\

\multicolumn{3}{c}{\textbf{AA (\%)}}
& 64.81±0.79
& 80.46±1.22
& 77.32±1.51
& 80.11±0.91
& 82.48±\std{0.46}
& 73.45±1.37
& 80.44±0.83
& 77.85±0.98
& 78.35±1.30
& 74.94±23.31
& 76.77±\std{0.46}
& \best{86.27±0.94} \\

\multicolumn{3}{c}{\textbf{$\kappa \times 100$ (\%)}}
& 34.39±1.53
& 59.97±1.87
& 51.30±2.01
& 54.52±3.23
& 59.73±2.44
& 43.65±2.44
& 54.82±1.54
& 44.94±2.78
& 56.43±2.55
& 54.08±18.28
& 40.26±\std{1.15}
& \best{65.28±2.84} \\

\bottomrule
\end{tabular}
}

\end{table*}
\subsection{Experimental Settings}

\subsubsection{Implementation and Training Details}

Our proposed STMamba is implemented on the PyTorch 2.1.0 platform using an NVIDIA RTX A6000 Ada Generation with 48GB of VRAM. The number of training epochs is 200. The Adam optimizer is adopted with a fixed learning rate \(\eta\) of 0.003 and weight decay 0.0001. The embedding dimension \(D\) is 64. The unified parameter configuration is used for all datasets. Mixed precision training is used.

\subsubsection{Comparing with Other Methods}
To compare the effectiveness of the proposed STMamba, we compare it against multiple SOTA HSI classification approaches from both quantitative and qualitative perspectives. The selected baselines cover various model architectures, including a traditional model: Random forest (RF), one CNN-based model:S\(^2\)VNet \cite{han2025subpixel}, one Transformer-based model: CPFormer \cite{11382038}, six Mamba-based models: SDMamba \cite{xu2025sparse}, MambaLG \cite{pan2025multiscale}, MambaHSI \cite{10604894}, MambaHSI+ \cite{wang2025mambahsiplus}, S\(^2\)Mamba \cite{s2mamba}, DSCC \cite{liu2026hyperspectral}, and two clustering-based methods: PSFormer \cite{10695122}, and DMSGer \cite{9927317}. Note that MambaLG, MambaHSI, MambaHSI+, and DSCC are patch-free Mamba models, while SDMamba and S\(^2\)Mamba are patch-based models and SDMamba has dynamic token selection based on feature similarity with a center anchor. MambaHSI+ and S\(^2\)Mamba perform bi-directional scanning, while SDMamba and MambaHSI perform single-direction scanning. MambaHSI, MambaHSI+, and S\(^2\)Mamba define the token at the regular pixel level, while DSCC defines the token on the superpixel level, named as supertoken. We report the mean value with standard deviation under 10 independent experiments.

\subsection{Experimental Results}
\subsubsection{Numerical Evaluation}

Quantitative results in terms of OA, AA, \(\kappa\), and per-class accuracy on PU, HU13, and HU18 are reported in \autoref{tab:pavia_results}, \autoref{tab:hu13_results}, and \autoref{tab:hu18_results}, respectively. The best results are highlighted in the tables.

\paragraph{PU}
As shown in \autoref{tab:pavia_results}, STMamba achieves the best overall performance, improving OA, AA, and \(\kappa\) over MambaHSI+ by 5.88\%, 5.90\%, and 7.17\%, respectively. MambaHSI+ outperforms MambaHSI, benefiting from bidirectional scanning, whereas S\(^{2}\)Mamba remains limited by patch-based processing, which restricts long-range dependency modeling. DSCC also adopts sparse semantic representations and achieves higher OA than SDMamba by exploiting complete supertoken regions, but its homogeneous region-level prediction results in lower AA. These comparisons demonstrate the advantage of constructing sparse semantic sequences while preserving fine-grained spatial information.

\paragraph{HU13}
As reported in \autoref{tab:hu13_results}, STMamba outperforms the second-best MambaLG by 0.97\%, 0.86\%, and 1.05\% in OA, AA, and \(\kappa\), respectively. MambaLG performs both within-patch and cross-patch scanning and therefore achieves stronger performance than most other Mamba-based baselines, highlighting the importance of modeling both local and global dependencies.

\paragraph{HU18}
On the larger HU18 scene, STMamba achieves the highest OA, AA, and \(\kappa\), exceeding MambaLG by 4.93, 3.79, and 5.55 percentage points, respectively. In contrast, SDMamba and S\(^{2}\)Mamba, which mainly perform directional scanning within fixed patches, achieve only 61.22\% and 50.77\% OA. MambaLG remains stronger than other conventional Mamba baselines, suggesting the importance of cross-scale and long-range modeling. The larger improvement of STMamba on HU18 further indicates the benefit of organizing spatially distant but semantically similar tokens into coherent sequences rather than relying on fixed spatial flattening.

\begin{figure}[!h]
\centering
\includegraphics[width=0.49\textwidth]{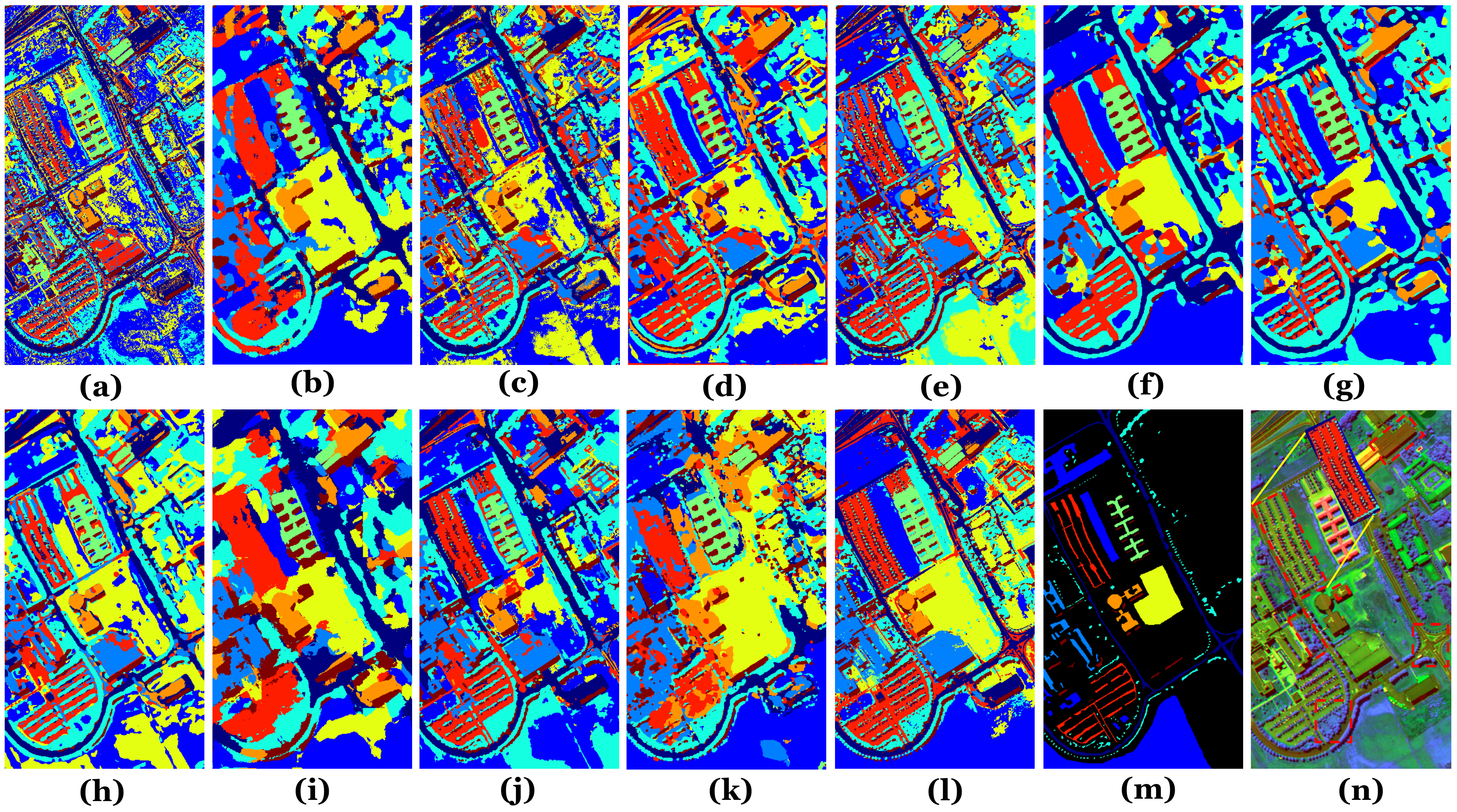}\\
\includegraphics[width=0.49\textwidth]{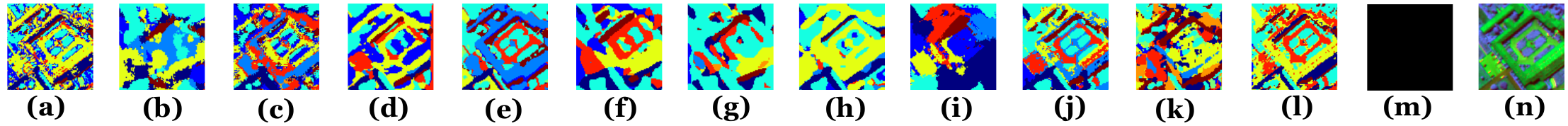}
\caption{Visual comparison on the PU dataset. (a) Random Forest, (b) CPFormer, (c) S\(^{2}\)VNet, (d) SDMamba, (e) MambaLG, (f) MambaHSI, (g) MambaHSI+, (h) S\(^{2}\)Mamba, (i) DSCC, (j) PSFormer, (k) DMSGer, (l) Ours, (m) Ground Truth, and (n) ICA Map.}
\label{fig:pavia_qualitative_comparison}
\end{figure}

\begin{figure}[!h]
\centering
\includegraphics[width=0.49\textwidth]{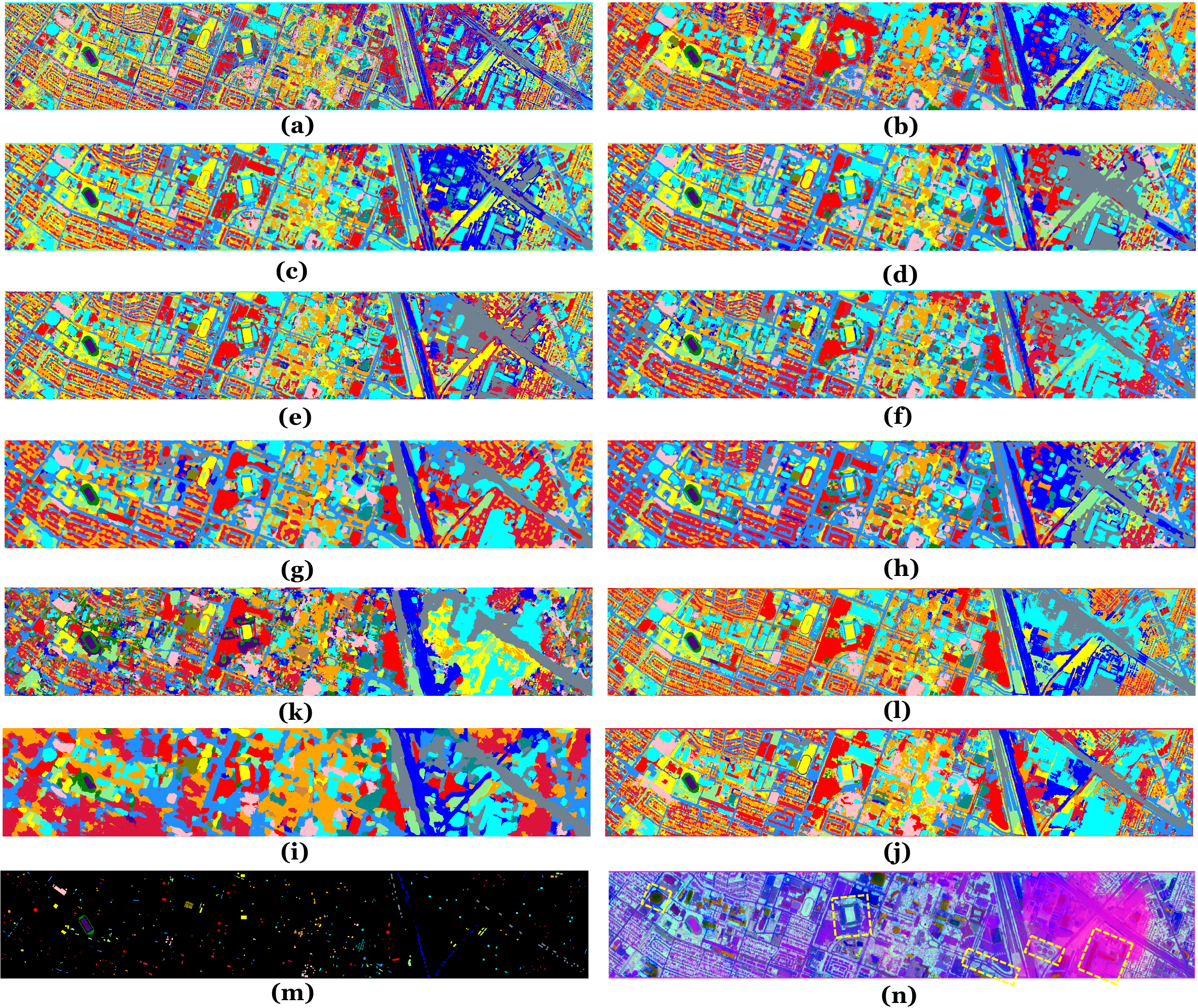}\\
\includegraphics[width=0.49\textwidth]{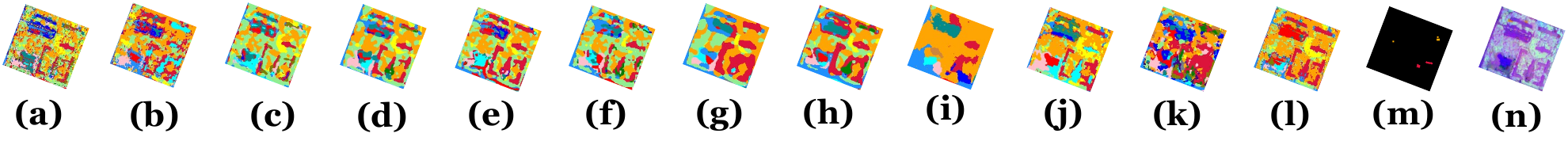}
\caption{Visual comparison on the HU13 dataset. (a) Random Forest, (b) CPFormer, (c) S\(^{2}\)VNet, (d) SDMamba, (e) MambaLG, (f) MambaHSI, (g) MambaHSI+, (h) S\(^{2}\)Mamba, (i) DSCC, (j) PSFormer, (k) DMSGer, (l) Ours, (m) Ground Truth, and (n) ICA Map.}
\label{fig:HU13_qualitative_comparison}
\end{figure}

\subsubsection{Visual Evaluation}

\begin{figure}[t]
\centering
\includegraphics[width=0.49\textwidth]{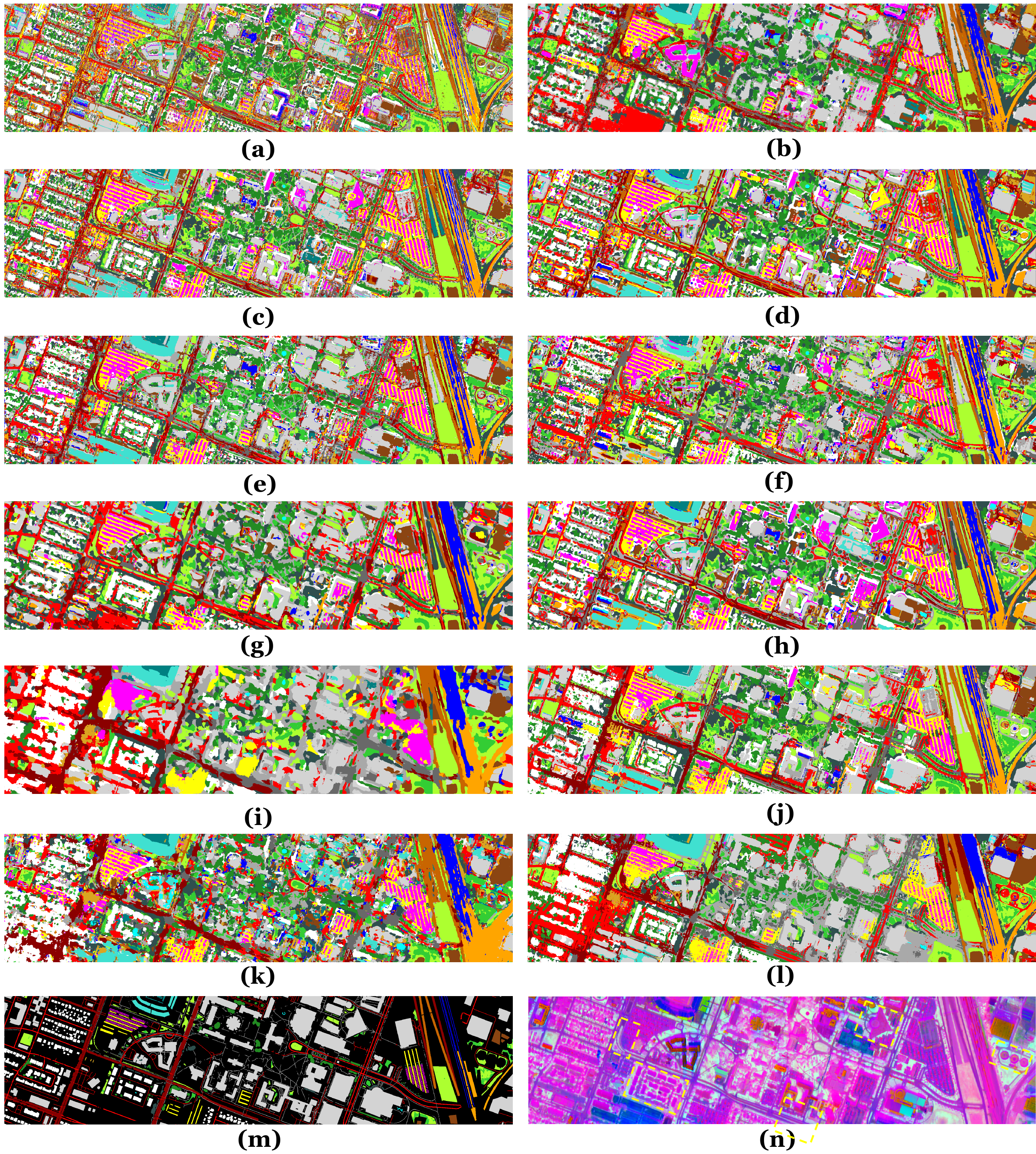}\\
\includegraphics[width=0.49\textwidth]{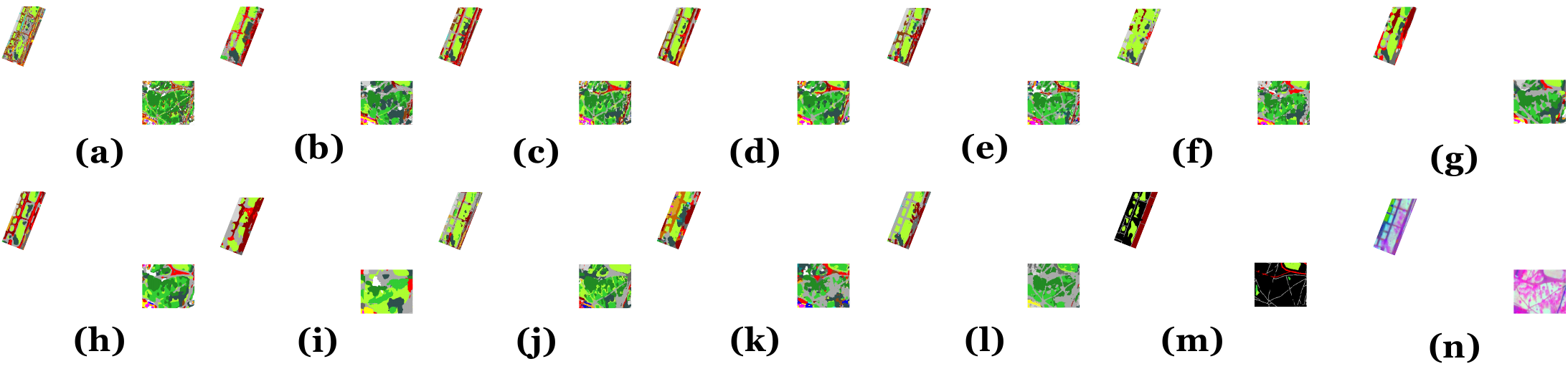}
\caption{Visual comparison on the HU18 dataset. (a) Random Forest, (b) CPFormer, (c) S\(^{2}\)VNet, (d) SDMamba, (e) MambaLG, (f) MambaHSI, (g) MambaHSI+, (h) S\(^{2}\)Mamba, (i) DSCC, (j) PSFormer, (k) DMSGer, (l) Ours, (m) Ground Truth, and (n) ICA Map.}
\label{fig:HU18_qualitative_comparison}
\end{figure}

The qualitative classification results are shown in \autoref{fig:pavia_qualitative_comparison}, \autoref{fig:HU13_qualitative_comparison}, and \autoref{fig:HU18_qualitative_comparison}. As shown in \autoref{fig:pavia_qualitative_comparison}, for PU dataset, STMamba preserves finer spatial details and object boundaries than the comparison methods. In particular, the road intersection, shadowed regions beneath tree canopies, building footprints, and the square grassland area are more clearly retained, while several baselines oversmooth these structures. The zoomed regions further show that STMamba better preserves small objects, including individual trees. In \autoref{fig:HU13_qualitative_comparison}, for HU13 dataset, STMamba better preserves linear, rectangular, and circular structures with clearer boundaries. For HU18 dataset, 
STMamba preserves more complete building structures and linear features than the comparison methods. S\(^{2}\)Mamba misclassifies some non-residential building regions as cars, while DMSGer produces more fragmented predictions within spatially continuous objects. The zoomed regions show that STMamba follows object boundaries more closely.

\subsubsection{Visualization of Clustering Center and Semantic Sequences}

\begin{figure}[t]
    \centering

    \subfloat[Stage 1]{
        \includegraphics[width=0.98\linewidth]{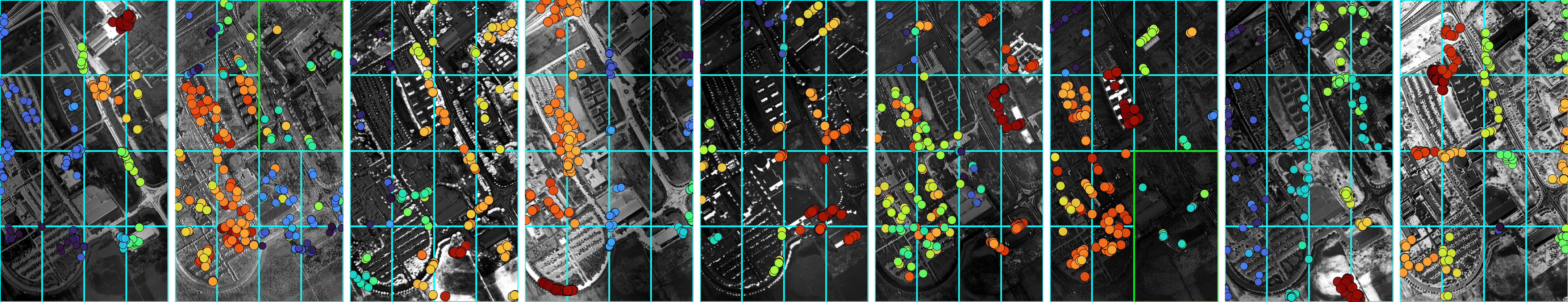}
        \label{fig:token_stage1}
    }\\[1mm]

    \subfloat[Stage 2]{
        \includegraphics[width=0.98\linewidth]{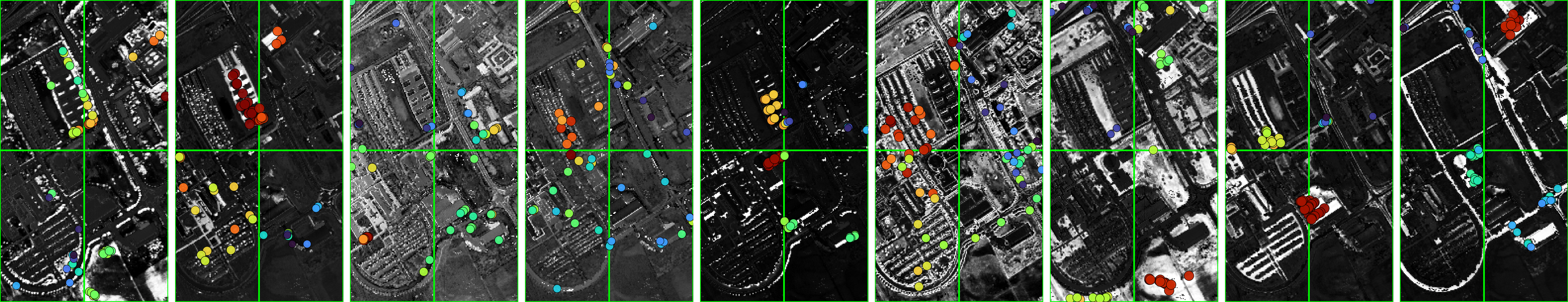}
        \label{fig:token_stage2}
    }\\[1mm]

    \subfloat[Stage 3]{
        \includegraphics[width=0.98\linewidth]{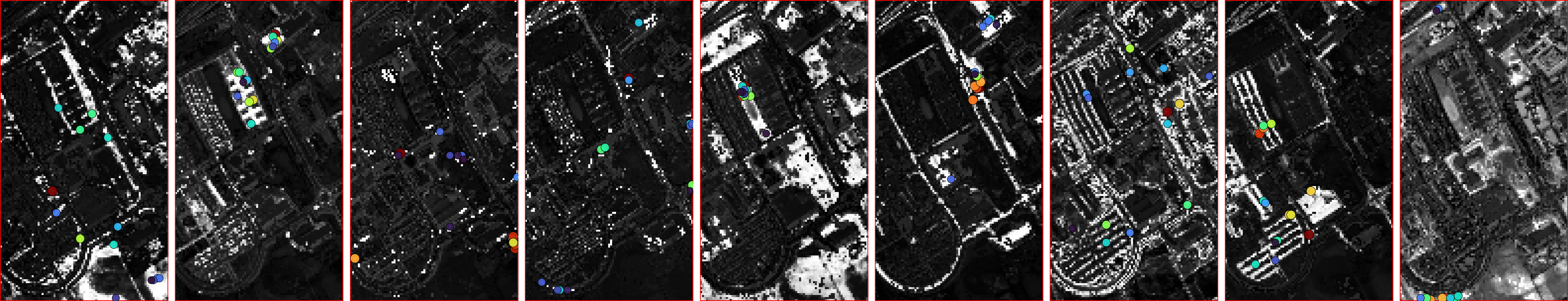}
        \label{fig:token_stage3}
    }\\[1mm]

    \subfloat[Stage 4]{
        \includegraphics[width=0.98\linewidth]{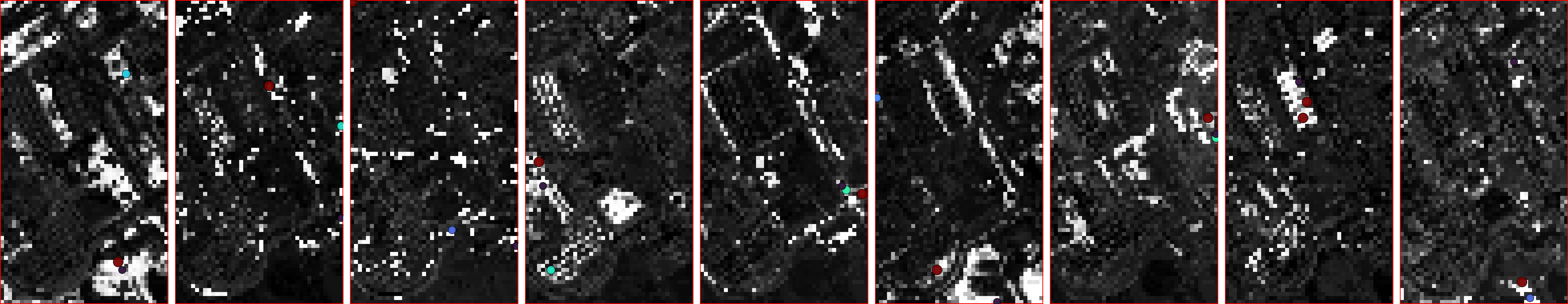}
        \label{fig:token_stage4}
    }

    \caption{Selected semantic tokens and their scanning order across four hierarchical stages.}
    \label{fig:selected_tokens}
\end{figure}

\begin{figure}[!h]
\centering
\includegraphics[width=0.49\textwidth]{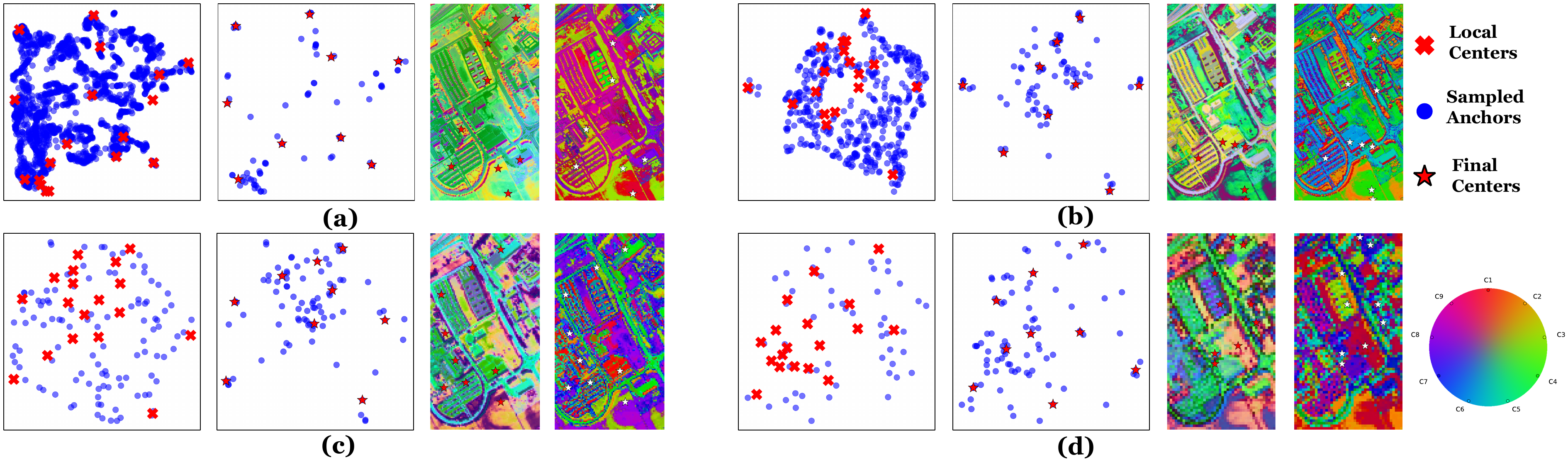}
\caption{Visualization of clustering centers across four STMamba stages on PU, including sampled anchors, aggregated centers, their spatial distributions, and semantic memberships.}
\label{fig:centers}
\end{figure}

As shown in \autoref{fig:selected_tokens}, semantic token selection becomes progressively more concentrated across stages as deeper features form clearer membership regions. These high-membership regions are then used to construct the semantic-topology sequences. Due to the sparsity control, \(\lambda_s\), latter stage will have fewer selected tokens. \autoref{fig:centers} shows that multi-sampling and the diversity loss $\mathcal{L}_{\mathrm{div}}$ produce more dispersed and separable cluster centers in feature space, while the corresponding centers remain sparsely distributed in image space and lie in semantically coherent membership regions.

\section{Discussion}
\label{discussion}

\subsection{Ablation Study}

\begin{figure}[]
\centering
\includegraphics[width=0.49\textwidth]{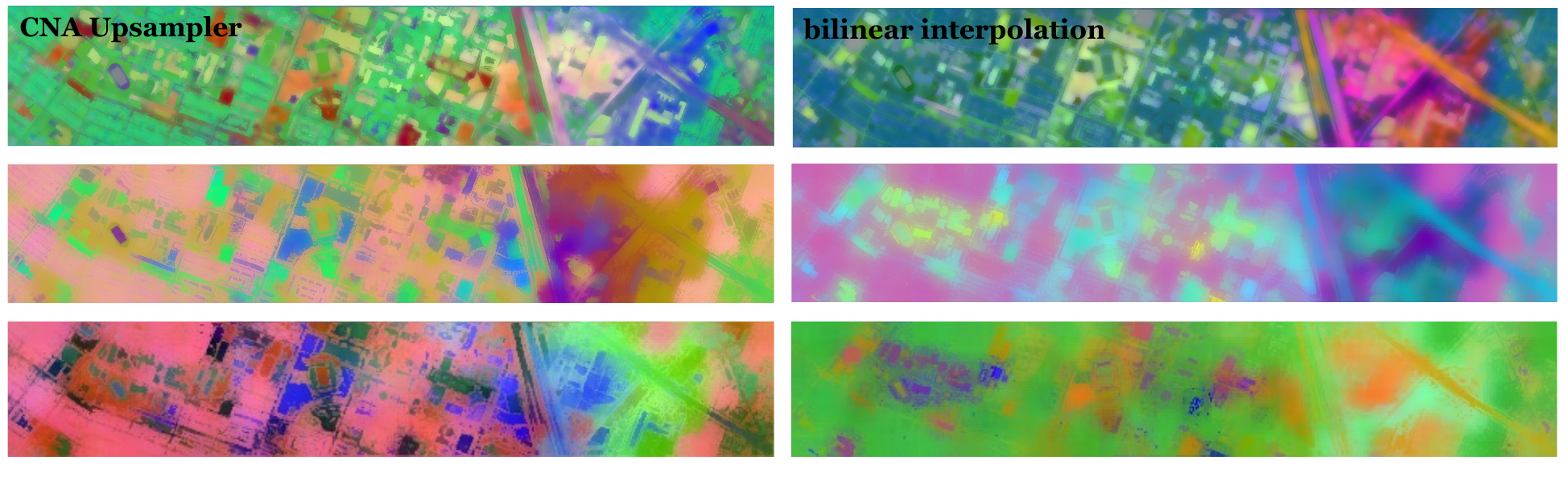}
\caption{PCA visualization of decoder features on the HU13 dataset by using CNA upsampler (left) and  bilinear-interpolation (right).}
\label{fig:upsampler_hu13}
\end{figure}

\autoref{tab:component_ablation} evaluates the major components of STMamba. Removing positional encoding, quadtree token selection, CNA, or the cluster-diversity loss consistently degrades performance, confirming their complementary contributions. In particular, replacing CNA with bilinear interpolation causes the largest degradation on HU18, reducing  OA from 71.34\% to 66.64\%, which highlights the importance of
cross-scale guidance for recovering dense spatial details from sparse features. Removing the complete SWSM pathway also substantially reduces performance, e.g., OA decreases from 95.97\% to 92.42\% on HU13,
demonstrating the effectiveness of semantic-wise sequence modeling. The visualizations in \autoref{fig:upsampler_hu13} shows that CNA produces clear structures than conventional interpolation.

\autoref{tab:SWSM_ablation} further analyzes the internal design of
SWSM. Removing Spa-SWSM consistently degrades performance, while
Spe-SWSM provides complementary spectral modeling, with the largest OA
drop of 2.98 percentage points on PU. Replacing membership-score
ordering with spatial ordering also reduces accuracy across all datasets,
showing the advantage of semantic ordering over geometric sequencing.
Similarly, removing membership gating degrades performance, particularly
on HU18, supporting membership-conditioned state propagation.

We evaluate the major components of STMamba on PU, HU13, and HU18 over 10 independent runs. The full model employs positional encoding, SWSM, quadtree token selection, the CNA upsampler, and the cluster-diversity loss, with $\lambda_s=0.01$ and $D=64$.


\begin{table}[h]
\centering
\caption{Ablation Study of the Main Components of the Proposed STMamba on Three Datasets.}
\label{tab:component_ablation}
\resizebox{0.49\textwidth}{!}{
\begin{tabular}{lccccc|ccc|ccc|ccc}
\toprule
\multirow{2}{*}{Setting} &
\multirow{2}{*}{Pos.} &
\multirow{2}{*}{SWSM} &
\multirow{2}{*}{Quad.} &
\multirow{2}{*}{CNA} &
\multirow{2}{*}{$\mathcal{L}_{div}$} &
\multicolumn{3}{c|}{PU} &
\multicolumn{3}{c|}{HU13} &
\multicolumn{3}{c}{HU18} \\
\cmidrule(lr){7-9}
\cmidrule(lr){10-12}
\cmidrule(lr){13-15}

& & & & & &
OA & AA & $\kappa$ &
OA & AA & $\kappa$ &
OA & AA & $\kappa$ \\
\midrule

w/o Pos.
& $\times$
& \checkmark
& \checkmark
& \checkmark
& \checkmark
& 84.10 & 93.57 & 80.17
& 95.77 & 96.46 & 95.42
& 71.14 & 85.94 & 65.04 \\

w/o SWSM
& $\times$
& $\times$
& \checkmark
& \checkmark
& $\times$
& 80.80 & 89.08 & 76.10
& 92.42 & 93.32 & 91.80
& 70.02 & 85.55 & 63.88 \\

w/o Quad-Tree
& \checkmark
& \checkmark
& $\times$
& \checkmark
& \checkmark
& 85.94 & 93.85 & 82.36
& 95.30 & 96.06 & 94.92
& 70.65 & 85.73 & 64.47 \\

w/o CNA
& \checkmark
& \checkmark
& \checkmark
& $\times$
& \checkmark
& 83.57 & 88.95 & 79.22
& 95.03 & 95.92 & 94.63
& 66.64 & 84.62 & 60.23 \\

w/o $\mathcal{L}_{div}$
& \checkmark
& \checkmark
& \checkmark
& \checkmark
& $\times$
& 85.88 & 93.30 & 82.23
& 95.61 & 96.35 & 95.25
& 70.40 & 85.79 & 64.24 \\

\midrule

Full Model
& \checkmark
& \checkmark
& \checkmark
& \checkmark
& \checkmark
& \textbf{86.12} & \textbf{94.23} & \textbf{82.62}
& \textbf{95.97} & \textbf{96.68} & \textbf{95.65}
& \textbf{71.34} & \textbf{86.27} & \textbf{65.28} \\

\bottomrule
\end{tabular}
}
\end{table}

\begin{table}[!bt]
\centering
\caption{Sensitivity Analysis of  Hyperparameters.}
\label{tab:parameter_sensitivity}
\resizebox{0.49\textwidth}{!}{
\begin{tabular}{lc|ccc|ccc|ccc}
\toprule
\multirow{2}{*}{Hyperparameter} &
\multirow{2}{*}{Value} &
\multicolumn{3}{c|}{PU} &
\multicolumn{3}{c|}{HU13} &
\multicolumn{3}{c}{HU18} \\
\cmidrule(lr){3-5}
\cmidrule(lr){6-8}
\cmidrule(lr){9-11}

& &
OA & AA & $\kappa$ &
OA & AA & $\kappa$ &
OA & AA & $\kappa$ \\
\midrule

\multirow{5}{*}{$\lambda_s$}
& 0.01 & 86.49 & 93.26 & 82.91 & 95.57 & 96.31 & 95.21 & 71.44 & 86.73 & 65.49 \\
& 0.05 & 87.67 & \textbf{94.95} & 84.37 & 96.63 & 97.21 & 96.36 & \textbf{72.09} & \textbf{86.91} & \textbf{66.23} \\
& 0.10 & \textbf{88.59} & 94.65 & \textbf{85.47} & \textbf{97.11} & \textbf{97.65} &\textbf{96.88} & 70.44 & 86.19 & 64.32 \\
& 0.20 & 88.52 & 93.13 & 85.27 & 96.96 & 97.50 & 95.71 & 71.48 & 86.49 & 65.35 \\
& 0.30 & 88.28 & 93.87 & 85.01 & 95.89 & 96.69 & 95.56 & 71.97 & 86.45 & 66.11 \\

\midrule

\multirow{3}{*}{$D$}
& 32  & 86.19 & 93.80 & 82.56 & 95.20 & 96.05 & 94.80 & 67.85 & 85.32 & 61.64 \\
& 64  & \textbf{86.98} & \textbf{94.39} & \textbf{83.52} & \textbf{96.68} & \textbf{97.27} & \textbf{96.41} & \textbf{71.34} & \textbf{86.73} & \textbf{65.49} \\
& 128 & 80.00 & 88.38 & 75.01 & 96.22 & 96.77 & 95.92 &  71.27 & 85.38 & 65.48 \\

\bottomrule
\end{tabular}
}
\end{table}

\subsection{Hyperparameter Analysis}

We investigate the global sparse ratio $\lambda_s$ and hidden dimension $D$ in \autoref{tab:parameter_sensitivity}. Classification accuracy does not monotonically increase with the number of retained tokens. PU and HU13 perform best around $\lambda_s=0.10$, whereas HU18 achieves its strongest overall performance at $\lambda_s=0.05$. Notably, the highly sparse setting $\lambda_s=0.01$ remains competitive, showing that a small subset of representative semantic tokens can preserve most discriminative information.

Token sparsity also directly affects memory consumption. Increasing $\lambda_s$ from 0.01 to 0.30 raises peak GPU memory from 2.69 to 5.04 GB on PU. Thus, $\lambda_s$ provides an explicit trade-off between accuracy and computational cost. For the hidden dimension, $D=64$ gives the most consistent performance across all three datasets.

\subsection{Analysis of Semantic Clustering}
\label{score emap}

\begin{figure}[h]
\centering
\includegraphics[width=0.49\textwidth]{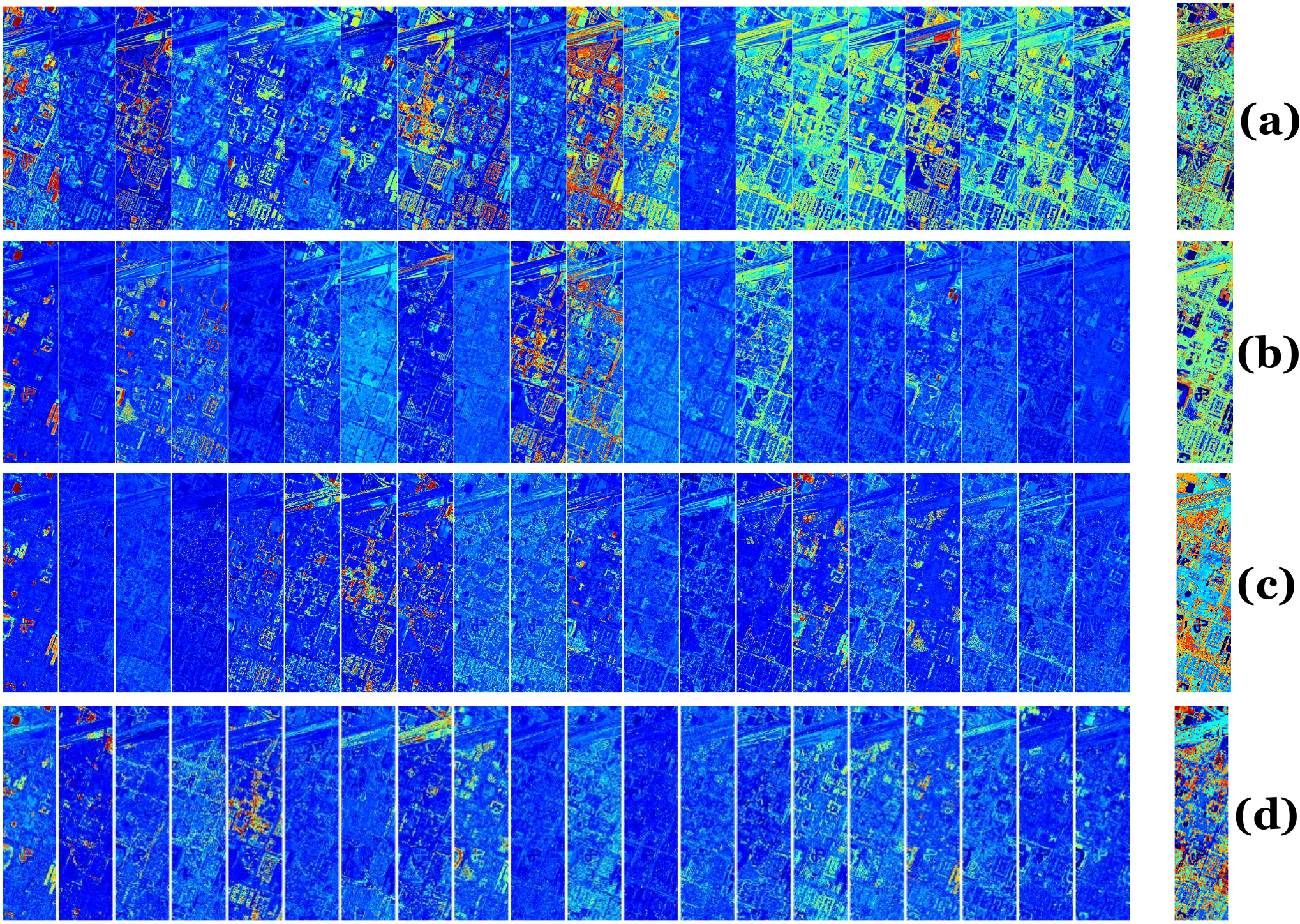}
\caption{Visualization of clustering membership and hard clustering map on the HU18 dataset at (a) stage1, (b) stage2, (c) stage3, and (d) stage4.}
\label{fig:clusteringmaps_HU18}
\end{figure}

The learned clustering centers and membership maps of HU18 are visualized in \autoref{fig:clusteringmaps_HU18}. Across hierarchical stages, the membership maps generally become more concentrated and semantically structured, while deeper-stage hard clustering maps show increasingly discriminative regions. Linear and block-shaped structures such as roads and buildings are clearly represented in multiple semantic memberships.

Separation is less distinct in some regions of the larger Houston scenes, potentially because of spectral similarity among classes or imperfect anchor sampling. Nevertheless, the learned memberships provide spatially coherent semantic groups that guide both sparse token selection and sequence ordering, rather than relying on predefined raster topology.


\newcommand{\diff}[1]{\,{\scriptsize\textcolor{gray}{(#1)}}}

\begin{table*}[htbp]
\centering
\caption{Ablation study of the internal designs of SWSM. 
Values in parentheses indicate the performance difference relative to the full model.}
\label{tab:SWSM_ablation}
\resizebox{\textwidth}{!}{
\small
\setlength{\tabcolsep}{4.2pt}
\renewcommand{\arraystretch}{1.15}

\begin{tabular}{lcccc ccc ccc ccc}
\toprule
\multirow{2}{*}{Setting} &
\multirow{2}{*}{\makecell{Spa-\\SWSM}} &
\multirow{2}{*}{\makecell{Spe-\\SWSM}} &
\multirow{2}{*}{\makecell{Membership Score \\Order Scan}} &
\multirow{2}{*}{\makecell{Membership\\Gate}} &
\multicolumn{3}{c}{PU} &
\multicolumn{3}{c}{HU13} &
\multicolumn{3}{c}{HU18} \\

\cmidrule(lr){6-8}
\cmidrule(lr){9-11}
\cmidrule(lr){12-14}

& & & & &
OA & AA & $\kappa$ &
OA & AA & $\kappa$ &
OA & AA & $\kappa$ \\

\midrule

w/o Spa-SWSM
& $\times$ & $\checkmark$ & $\checkmark$ & $\checkmark$
& 84.72\diff{-2.26}
& 93.76\diff{-0.20}
& 80.96\diff{-2.61}
& 95.13\diff{-1.00}
& 95.90\diff{-0.83}
& 94.74\diff{-1.07}
& 68.80\diff{-2.60}
& 84.91\diff{-1.18}
& 62.55\diff{-2.74} \\

w/o Spe-SWSM
& $\checkmark$ & $\times$ & $\checkmark$ & $\checkmark$
& 84.00\diff{-2.98}
& 93.45\diff{-0.51}
& 80.13\diff{-3.44}
& 95.27\diff{-0.86}
& 95.96\diff{-0.77}
& 94.88\diff{-0.93}
& 70.95\diff{-0.45}
& 86.09\diff{0.00}
& 64.89\diff{-0.40} \\

Spatial-order Scan
& $\checkmark$ & $\checkmark$ & $\times$ & $\checkmark$
& 85.57\diff{-1.41}
& 93.74\diff{-0.22}
& 81.92\diff{-1.65}
& 95.42\diff{-0.71}
& 96.28\diff{-0.45}
& 95.05\diff{-0.76}
& 70.43\diff{-0.97}
& 85.49\diff{-0.60}
& 64.25\diff{-1.04} \\

w/o Membership Gate
& $\checkmark$ & $\checkmark$ & $\checkmark$ & $\times$
& 86.30\diff{-0.68}
& 93.76\diff{-0.20}
& 82.84\diff{-0.73}
& 95.66\diff{-0.47}
& 96.36\diff{-0.37}
& 95.30\diff{-0.51}
& 69.78\diff{-1.62}
& \textbf{86.18}\diff{+0.09}
& 63.66\diff{-1.63} \\

\midrule

\textbf{Full Model}
& $\checkmark$ & $\checkmark$ & $\checkmark$ & $\checkmark$
& \textbf{86.98}
& \textbf{93.96}
& \textbf{83.57}
& \textbf{96.13}
& \textbf{96.73}
& \textbf{95.81}
& \textbf{71.40}
& 86.09
& \textbf{65.29} \\

\bottomrule
\end{tabular}
}
\end{table*}



\begin{figure}[h]
    \centering

    \subfloat[Overall Accuracy (OA)]{
        \includegraphics[width=\linewidth]
        {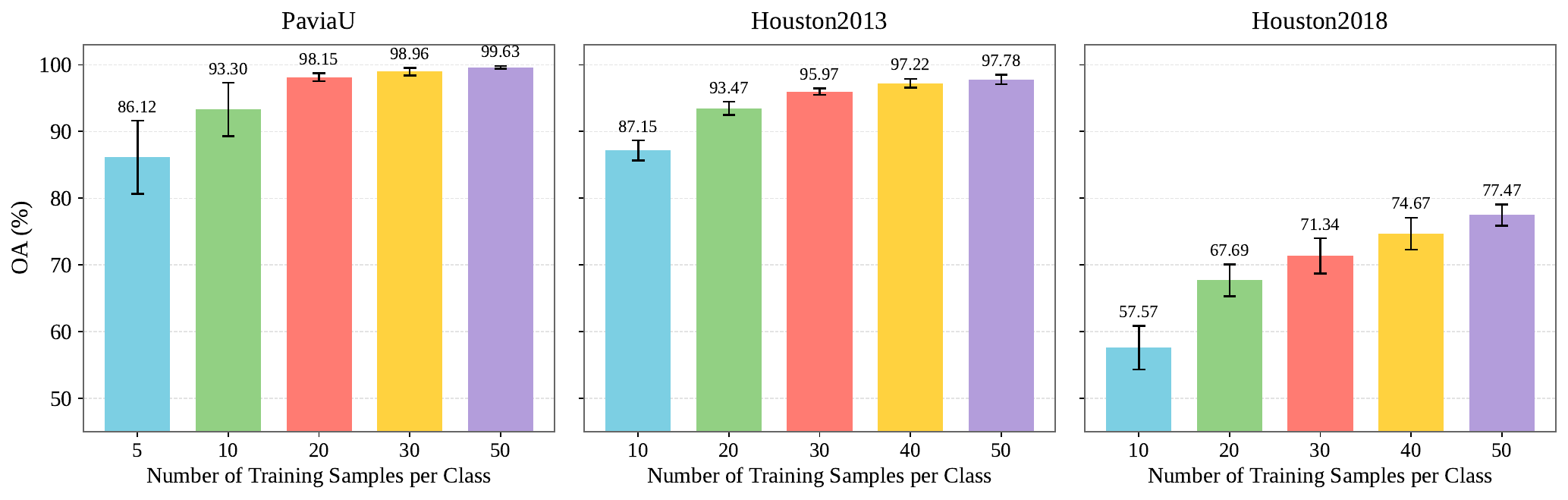}
        \label{fig:train_num_oa}
    }



    \caption{Performance of the proposed method under different numbers of training samples per class on the PU, HU13, and HU18 datasets. Results are reported over 10 independent runs.}
    \label{fig:different_training_samples}
\end{figure}

\subsection{Effect of Training Samples}


As shown in \autoref{fig:different_training_samples}, OA consistently improves with more labeled samples and gradually saturates on PU and HU13, whereas HU18 benefits more from additional supervision.

\subsection{Computational Complexity}

\autoref{tab:complexity} compares STMamba with patch-free Mamba-based HSIC methods. STMamba contains approximately 0.6M parameters across all three datasets, substantially fewer than DSCC and MambaHSI+ and comparable to lightweight Mamba baselines. Although its GFLOPs are higher than MambaHSI and MambaHSI+ because of hierarchical feature processing and FFNs, STMamba remains considerably less computationally expensive than MambaLG on the large Houston scenes. These results indicate a favorable balance between model capacity, computational cost, and global semantic modeling.

\begin{table}[!t]
\centering
\caption{Comparison of model parameters and computational complexity.}
\label{tab:complexity}
\setlength{\tabcolsep}{3.2pt}
\renewcommand{\arraystretch}{1.12}

\resizebox{\columnwidth}{!}{
\begin{tabular}{lcc cc cc}
\toprule
\multirow{2}{*}{Method}
& \multicolumn{2}{c}{PU}
& \multicolumn{2}{c}{HU13}
& \multicolumn{2}{c}{HU18} \\
\cmidrule(lr){2-3}
\cmidrule(lr){4-5}
\cmidrule(lr){6-7}

& Params (M) & GFLOPs
& Params (M) & GFLOPs
& Params (M) & GFLOPs \\
\midrule

MambaLG
& 0.293 & 80.273
& 0.297 & 262.881
& 0.289 & 543.230 \\

MambaHSI
& 0.412 & 5.938
& 0.418 & 26.078
& 0.407 & 21.108 \\

MambaHSI+
& 1.631 & 9.703
& 1.643 & 38.652
& 1.636 & 49.168 \\

DSCC
& 24.580 & 73.570
& 24.616 & 76.682
& 24.537 & 69.397 \\

\midrule

Ours
& 0.602 & 31.952
& 0.605 & 106.367
& 0.599 & 212.674 \\

\bottomrule
\end{tabular}}
\end{table}

\begin{figure}[]
\centering
\includegraphics[width=0.49\textwidth]{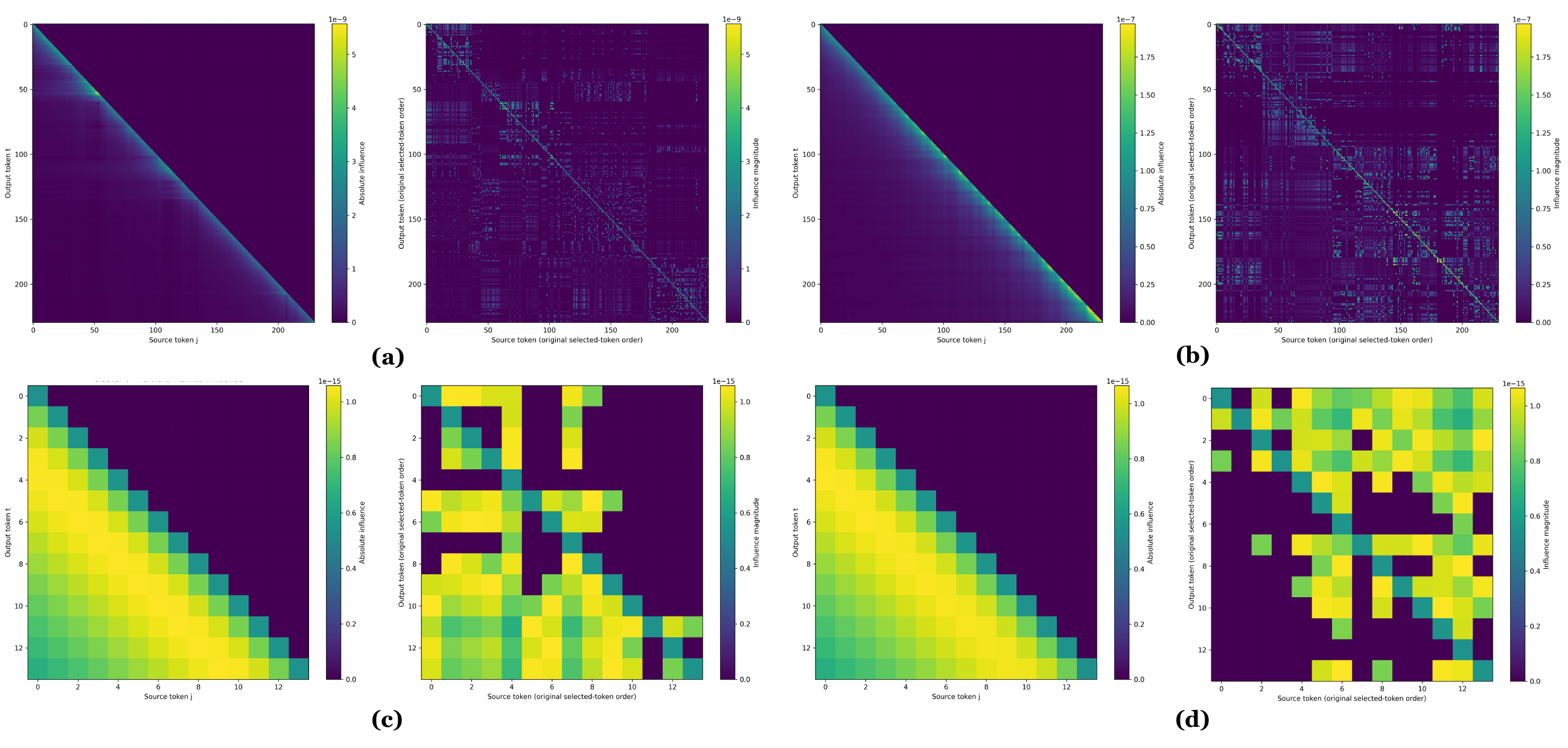}
\caption{
Visualization of the hidden influence matrices in SWSM.
At stage 1, (a) shows the forward semantic-topology scanning influence matrix
(left) and the corresponding permutation-restored token influence matrix
(right) for cluster 1, while (b) shows the same visualization for cluster 8.
At stage 3, (c) and (d) present the corresponding results for clusters 1 and 8,
respectively.
}
\label{fig:attentionmatrix}
\end{figure}

\subsection{Visualization of Hidden Influence Matrices in Mamba}
\label{decay}

To analyze state propagation in SWSM, \autoref{fig:attentionmatrix} visualizes the hidden influence matrices derived from \autoref{eq:s6_matrix_expanded}. The influence generally decreases as sequence distance increases, consistent with the recurrent transition term
$\prod_{k=j+1}^{i}\bar{\mathbf{A}}_k$.
This observation provides additional motivation for reducing sequence redundancy and organizing representative tokens into compact semantic sequences.

Because SWSM reorders tokens before scanning, the influence matrix is initially defined in semantic-sequence space. We therefore restore it to the original selected-token topology as
\begin{equation}
\mathbf{A}_{\mathrm{PRTIM}}
=
\mathbf{T}^{\top}
\mathbf{P}^{\top}
\mathbf{A}
\mathbf{P}
\mathbf{T},
\end{equation}
where $\mathbf{A}$ is the influence matrix in semantic scanning order, while $\mathbf{P}$ and $\mathbf{T}$ denote permutation and token-organization operations. The restored matrix reveals the effective interactions induced among the originally selected tokens.


\section{Conclusion} \label{conclusion}



This paper presented STMamba, a hierarchical framework that constructs sparse semantic-token sequences through token clustering and models them using spatial and spectral SWSM. Density-aware center discovery, quadtree-based token selection, and membership-conditioned sequencing enable long-range modeling among semantically coherent tokens, while CNA progressively restores dense spatial details. Experiments on three large-scale HSI datasets demonstrate competitive classification performance and improved preservation of fine structures. Current limitations include the fixed cluster number and additional cost of quadtree selection; future work will investigate adaptive clustering and more efficient dynamic sequence construction.








%
\bibliographystyle{IEEEtran}
\bibliography{IEEEabrv,ref}

%








\end{document}